\documentclass[sigconf]{acmart}

\AtBeginDocument{%
  }

\usepackage{multirow}
\usepackage{enumitem}
\usepackage{makecell}
\usepackage{ltxtable}
\usepackage{listings}
\renewcommand\footnotetextcopyrightpermission[1]{} % removes footnote with conference information in first column
\begin{document}

%%
%% The "title" command has an optional parameter,
%% allowing the author to define a "short title" to be used in page headers.
\title{The Vision of Autonomic Computing: Can LLMs Make It a Reality?}

%%
%% The "author" command and its associated commands are used to define
%% the authors and their affiliations.
%% Of note is the shared affiliation of the first two authors, and the
%% "authornote" and "authornotemark" commands
%% used to denote shared contribution to the research.
% \author{Ben Trovato}
% \authornote{Both authors contributed equally to this research.}
% \email{trovato@corporation.com}
% \orcid{1234-5678-9012}
% \author{G.K.M. Tobin}
% \authornotemark[1]
% \email{webmaster@marysville-ohio.com}
% \affiliation{%
%   \institution{Institute for Clarity in Documentation}
%   \city{Dublin}
%   \state{Ohio}
%   \country{USA}
% }

% \author{Lars Th{\o}rv{\"a}ld}
% \affiliation{%
%   \institution{The Th{\o}rv{\"a}ld Group}
%   \city{Hekla}
%   \country{Iceland}}
% \email{larst@affiliation.org}

%% The "author" command and its associated commands are used to define
%% the authors and their affiliations.

\author{Zhiyang Zhang $^1$\footnotemark[1] \quad Fangkai Yang $^2$ \quad Xiaoting Qin $^2$ \quad Jue Zhang $^2$\footnotemark[2] \quad Qingwei Lin $^2$ \\
    Gong Cheng $^1$ \quad Dongmei Zhang $^2$ \quad Saravan Rajmohan $^2$ \quad Qi Zhang $^2$ \\
    $^1$State Key Laboratory for Novel Software Technology, Nanjing University \\
    $^2$Microsoft}

%%
%% By default, the full list of authors will be used in the page
%% headers. Often, this list is too long, and will overlap
%% other information printed in the page headers. This command allows
%% the author to define a more concise list
%% of authors' names for this purpose.
% \renewcommand{\shortauthors}{Trovato et al.}

%%
%% The abstract is a short summary of the work to be presented in the
%% article.
\begin{abstract}

The Vision of Autonomic Computing (ACV), proposed over two decades ago, envisions computing systems that self-manage akin to biological organisms, adapting seamlessly to changing environments. Despite decades of research, achieving ACV remains challenging due to the dynamic and complex nature of modern computing systems. Recent advancements in Large Language Models (LLMs) offer promising solutions to these challenges by leveraging their extensive knowledge, language understanding, and task automation capabilities. This paper explores the feasibility of realizing ACV through an LLM-based multi-agent framework for microservice management. We introduce a five-level taxonomy for autonomous service maintenance and present an online evaluation benchmark based on the Sock Shop microservice demo project to assess our framework's performance. Our findings demonstrate significant progress towards achieving Level 3 autonomy, highlighting the effectiveness of LLMs in detecting and resolving issues within microservice architectures. This study contributes to advancing autonomic computing by pioneering the integration of LLMs into microservice management frameworks, paving the way for more adaptive and self-managing computing systems. The code will be made available at \url{https://aka.ms/ACV-LLM}.
\end{abstract}

\maketitle
\renewcommand{\shortauthors}{Zhang et al.}
\pagestyle{plain}

\renewcommand{\thefootnote}{\fnsymbol{footnote}}
\footnotetext[1]{Work is done during an internship at Microsoft.} 
\footnotetext[2]{Corresponding author.} 
\renewcommand{\thefootnote}{\arabic{footnote}}

\section{Introduction}

As we enter the modern era, computing infrastructure is becoming increasingly distributed and large-scale, presenting significant challenges for human management. This complexity underscores the necessity for developing systems capable of self-management. This vision, reminiscent of the Vision of Autonomic Computing (ACV)~\cite{Kephart2003TheVO} proposed two decades ago, envisions computing systems that can manage themselves according to an administrator's goals, integrating new components as seamlessly as biological cells. 

Realizing an autonomic system with self-management poses significant challenges. Early approaches to autonomic systems often relied on rule-based mechanisms and predefined policies, which, while effective to some extent, struggled to adapt to the increasingly dynamic and complex environments seen in modern computing systems \cite{Huebscher2008survey, Lalanda2013survey}. Over the years, significant strides have been made towards achieving this vision through the development of self-adaptive and self-managing systems~\cite{Sukhpal2022survey, Li2024survey}. Despite significant efforts and progress over the past two decades, the realization of ACV is still elusive due to numerous grand challenges outlined in the ACV paper, many of which hinge on breakthroughs in AI.

Recent advancements in AI, particularly through Large Language Models (LLMs), offer promising new avenues to address these challenges. LLMs' extensive knowledge, language understanding, and task automation capabilities~\cite{brown2020language, ouyang2022training, openai2023gpt4} represent a significant leap forward in our ability to create truly autonomic systems. Successful demonstrations of LLMs in tasks such as anomaly detection and incident mitigation \cite{zhang2024survey} illustrate their potential to provide the contextual understanding and adaptive decision-making necessary to achieve the goals of ACV. This prompts an intriguing question: \textbf{Can LLMs make the Vision of Autonomic Computing a reality?} 

Addressing this question is complex due to the broad scope of ACV. To systematically assess the feasibility of achieving ACV using LLMs, evaluation studies in concrete settings are necessary. This work proposes to study this feasibility within the context of microservice self-management, a popular architecture for managing cloud services. Specifically, we propose an LLM-based multi-agent self-management framework for microservices and assess its performance using a live online evaluation benchmark built on the known microservice demo project Sock Shop~\cite{sockshop}.

Our proposed LLM-based microservice management system employs a hierarchical multi-agent architecture. High-level group manager handles declarative tasks that span multiple service components, such as optimizing end-to-end latency to under 200 ms. In contrast, low-level autonomic agents focus on specific tasks within their managed service components. To evaluate our system, we introduce a five-level taxonomy of autonomous service maintenance, emphasizing Self-Optimization and Self-Healing. We then design specific evaluation tasks within the Sock Shop microservice, employing chaos engineering techniques to deliberately introduce faults and observe how our management system resolves these issues. Our findings demonstrate that \textbf{the LLM-based multi-agent framework achieves Level 3 autonomy in our five-level taxonomy}. While it effectively detects issues and performs specific imperative tasks, there are opportunities for further enhancements in root cause analysis and issue mitigation capabilities.

Our contributions can be summarized as follows: 
\begin{itemize} [noitemsep, left=0pt]
    \item We advance the domain of autonomic computing for microservice management through an LLM-based multi-agent framework. To the best of our knowledge, we are the first research work to explore microservice self-management using LLM-based agents.
    \item We establish a taxonomy consisting of five levels for autonomous service maintenance. We also present an online evaluation benchmark designed to assess tasks corresponding to each level of autonomy within the microservice demo project Sock Shop.
    \item We conduct a rigorous evaluation of our LLM-based microservice management framework using the aforementioned benchmark.
\end{itemize}

\section{Background and Related Work}

\noindent{\textbf{Autonomic Computing.}} 
The goal of autonomic computing is to develop self-managing systems which reduces the complexity and cost of IT management that increases system reliability and performance. It is inspired by the biological autonomic nervous system, which autonomously regulates functions such as heart rate and body temperature, thereby reducing cognitive load. The ACV paper identified four key objectives of self-management in autonomic systems~\cite{Kephart2003TheVO, Figueira2024mape}: 
\begin{itemize} [noitemsep, left=0pt]
    \item Self-Configuration: Systems can configure and re-adjust themselves to meet the agreed objectives.
    \item Self-Optimization: Continuously monitor themselves and seek opportunities to improve performance and costs.
    \item Self-Healing: The ability to autonomously recover from failures and even predict them.
    \item Self-Protection: Against malicious attacks or cascading failures.
\end{itemize}
It also proposed the Monitor, Analyze, Plan, Execute, and Knowledge (MAPE-K) loop~\cite{Kephart2003TheVO} as a structured approach to autonomic computing, providing a clear and systematic methodology for implementing self-managing capabilities. The MAPE-K model forms the foundation of self-adaptive and self-management systems, enabling continuous adaptation and optimization through feedback loops~\cite{ Arcaini2015ModelingAA, Weyns2018mape, Mendonca2021mape, Figueira2024mape} and creating autonomous systems to handle complex and dynamic environments.

Subsequent works have explored various aspects of autonomic systems, leading to significant advancements in self-adaptive systems, such as handling uncertainties and runtime variabilities~\cite{Incerto2017adapt, Gerasimou2018probabilistic, Zhao2021probabilistic, Aradea2023aras, Kienzle2023decision}, as well as self-protecting ability in terms of security risks~\cite{Hammad2018android}. Autonomic computing principles have also been applied beyond software systems, such as in robotics~\cite{Camilli2023adpt}, autonomous driving systems~\cite{Sagar2016drive}, and the development of digital twins~\cite{Guine2024dt}. These systems require sophisticated self-adaptation mechanisms to cope with dynamic and unpredictable environments.

Initially, autonomic systems often relied on rule-based approaches and predefined policies to achieve self-management. While effective in certain contexts, these approaches were limited in their ability to handle complex, dynamic scenarios requiring adaptive decision-making and contextual understanding. Recent advancements in AI and machine learning (ML) have introduced new possibilities. Leveraging AI and ML to provide deeper insights (such as event correlation and predictive analytics) and automation capabilities has further enhanced the self-managing capabilities of IT operations, leading to the development of a field known as AIOps.

\noindent{\textbf{Autonomic Management of Cloud-Native Applications.}} Cloud computing has emerged as a critical component of modern distributed computing systems, offering scalable resources, cost efficiency, and flexibility through on-demand access to computing power, storage, and applications. This shift has led to the widespread adoption of cloud-native applications, leveraging architectures such as microservices. However, managing these applications poses significant challenges, including ensuring security, navigating complex microservices architectures, maintaining observability, managing resource allocation, and ensuring reliable performance and scalability~\cite{Speth2022issue, Mendonca2021mape, Yang2022aid, Hemanth2022security}.

To address these challenges, numerous management tools have been developed. The Cloud Native Computing Foundation (CNCF)~\cite{cncf_website} hosts various projects aimed at fostering innovation and collaboration within the cloud-native community, such as Kubernetes~\cite{kubernetes_overview} and Prometheus~\cite{prometheus_website}. Kubernetes, for example, automates deployment, scaling, and management of containerized applications, offering features like service discovery and automated rollouts. Despite these capabilities, Kubernetes lacks comprehensive high-level management features aligned closely with user intent, such as built-in middleware or advanced configuration management systems. 
On the research side, recent advancements, such as the AMoCNA~\cite{Kosiska2020AutonomicMF, Kosiska2023EnhancementOC, Kosiska2023ExperimentalEO} framework, extend traditional autonomic management approaches like the MAPE-K loop~\cite{Arcaini2015ModelingAA} to enhance autonomy in cloud-native environments. However, these frameworks often rely on rule-based systems, limiting their adaptability in handling complex scenarios requiring contextual understanding and adaptive decision-making. To address these limitations, recent works have further extended the traditional MAPE-K loop to incorporate AI and ML advancements~\cite{Tundo2023energyadpapt, Nunes2024microservice}, proposing a novel self-adaptation approach~\cite{Kulkarni2023adapt, Vaidhyanathan2024ML}. 

\noindent\textbf{LLM-based Kubernetes and Cloud Service Management.} 
In the era of LLMs, integrating LLMs into Kubernetes management represents a promising avenue to overcome limitations in traditional approaches. For instance, GenKubeSec~\cite{malul2024genkubesecllmbasedkubernetesmisconfiguration}  applies LLM-based methods to detect and remediate Kubernetes configuration file misconfigurations with precision and detailed insights, surpassing the capabilities of rule-based tools. Similarly, the open-source project K8sGPT~\cite{k8sgpt_docs} provides natural language-based diagnostics and issue triaging for Kubernetes clusters. Despite these advancements, a cohesive integration of LLMs across all facets of self-management in Kubernetes remains an ongoing area of development.

Similarly, current cloud service management systems are starting to integrate LLMs for AIOps to ensure the high availability and reliability of large-scale cloud infrastructures~\cite{zhang2024survey,xu2024large,zhang2024llms}. The key AIOps tasks within cloud service management include data collection and pre-processing, root cause analysis (RCA), and incident mitigation.
Cloud platforms consist of numerous services with diverse configurations, posing challenges for traditional AIOps models in terms of cross-service and cross-task flexibility. LLMs offer a solution by efficiently interpreting unstructured data such as service logs and troubleshooting guides, extracting essential information from large data volumes~\cite{le2023log,jin2023assess}. Once an anomaly or incident is detected, RCA is crucial for identifying the underlying causes. Leveraging information extracted and summarized by LLM-based data pre-processing, LLMs can comprehend and localize incidents effectively~\cite{ahmed2023recommending,chen2024automatic}. Incident mitigation follows RCA, where LLMs assist in automating mitigation steps previously managed by on-call engineers (OCEs)~\cite{qiao2023taskweaver,an2024nissist}. These LLM-based solutions empower each AIOps task by reducing human effort and increasing automation. However, despite these advancements in each AIOps task, the entire AIOps process is not yet fully self-managed, requiring additional efforts for managing large-scale cloud service systems. The concrete realization of ACV in our paper offers a demonstration that unifies and coordinates LLM-based AIOps task solutions to achieve a self-managed cloud service system with high efficiency, availability and reliability.

\section{Service Management with LLM-based Multi-Agents}
\label{sec:framework}

In this section, we demonstrate the application of ACV in service management using LLM-based agents, realizing the general architectural considerations discussed in the ACV paper.

\begin{figure}[htb!]
  \includegraphics[width=0.9\columnwidth]{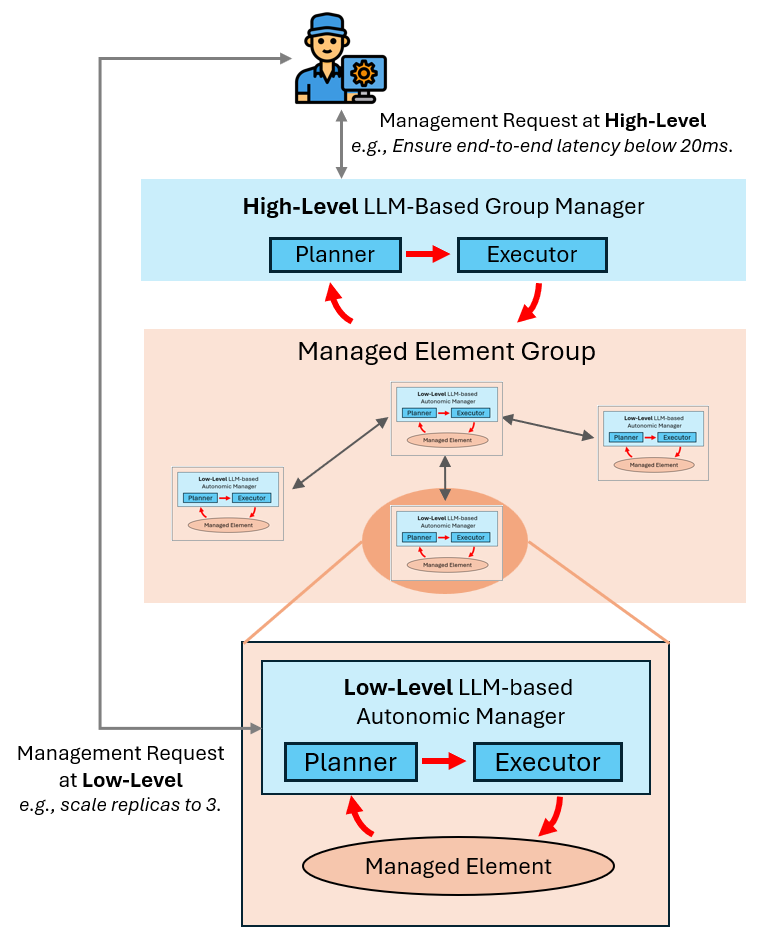}
  \caption{Hierarchical service management framework with LLM-based multi-agent design.}
  \label{fig:two-level-architecture}
  \vspace{-2mm}
\end{figure}

\begin{figure*}[htb!]
    \includegraphics[width=0.95\textwidth]{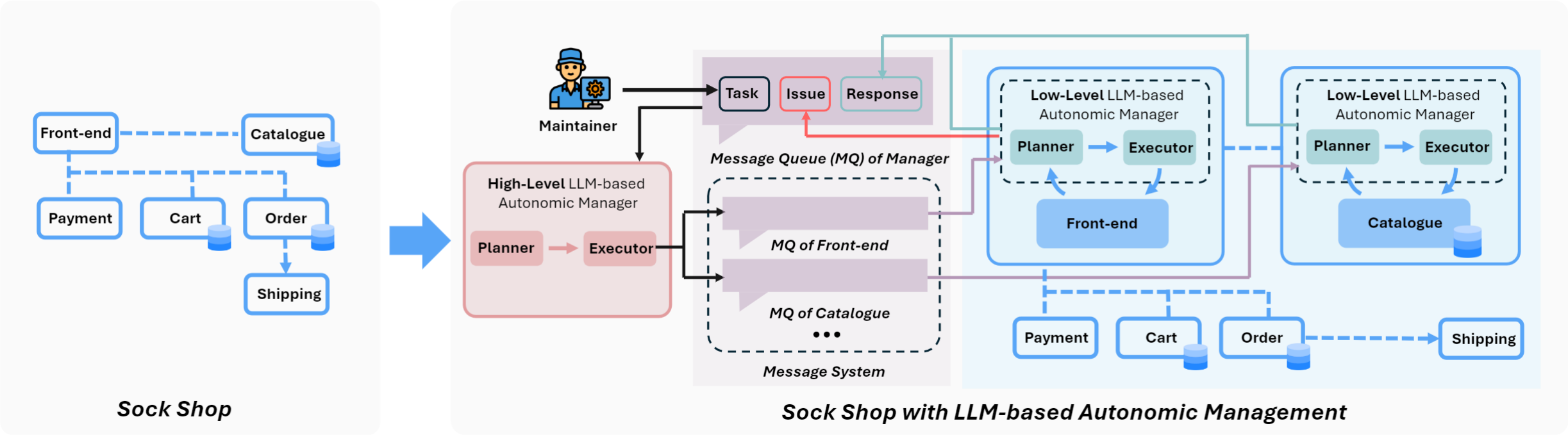}
  \caption{Illustration of the Sock Shop example: Each microservice is converted into an LLM-based autonomic agent, managed by an autonomic layer comprising a Planner and Executor. The high-level group manager employs a Plan-Execute feedback mechanism to generate plans and assign sub-tasks to low-level autonomic agents. A decoupled message queue system serves as the middleware to manage communication, collecting feedback and unresolved issues.}
  \label{fig:sock-shop-implementation}
\end{figure*}

\noindent\textbf{Architecture Overview.} The ACV paper emphasizes that autonomic management elements operate at various levels, from individual service components to service groups fulfilling specific features, and up to entire business functions. This concept aligns with the design of LLM-based multi-agent system~\cite{wang2024survey,guo2024large}, where a hierarchical architecture is naturally suited to manage services at these different levels. Considering the varied complexity of management requests at different levels, we propose a two-level autonomic management architecture in this paper. As shown in Figure~\ref{fig:two-level-architecture}, this hierarchical service management framework provides flexibility for the maintainers to manage the system at different appropriate levels. For complex management requests, such as ``\textit{ensure end-to-end latency below 20 ms}'', which involve multiple services, the high-level LLM-based group manager processes the request by breaking it down into simpler low-level tasks that can be handled by the respective low-level autonomic agents. Conversely, simpler requests can be sent directly to the low-level autonomic agents. For example, a request to ``\textit{scale replicas to 3}'' can be directed straight to the corresponding low-level autonomic agent, bypassing the high-level group manager. The detailed working mechanism of this two-level architecture will be discussed later in this section.

\noindent\textbf{Low-Level Autonomic Agent.} 
Traditionally, a system consists of elements with limited autonomic capabilities responsible for basic service functions, such as a pod in Kubernetes. These elements operate reactively, executing tasks like responding to API calls upon request. By integrating an autonomic layer to manage each element, these managed elements can enhance their functionality significantly. They can monitor their own health, analyze potential issues, and develop mitigation plans to resolve problems autonomously.

Before the advent of LLMs, autonomic capabilities were achieved through explicit development of Monitor, Analyze, Plan, Execute, and Knowledge modules~\cite{Kephart2003TheVO}. However, LLMs streamline these functions into two core modules: \textbf{Planner} and \textbf{Executor}. The Planner generates execution steps to achieve specific goals, including monitoring and analysis, 
% which may be integrated in system prompts or specified by user queries. 
which maybe passing down from the high-level group manager (discussed in the below section) or requested directly by maintainers.
These steps are often executable maintenance code (e.g., \texttt{kubectl} commands in Kubernetes), leveraging the coding capabilities and general knowledge of LLMs. The Executor carries out these steps, and the results are sent back to the Planner to verify goal achievement. 
If the goal is unmet, another self-correction cycle is initiated. This Plan-Execute feedback loop simplifies the traditional design of self-management agents, facilitating the development of autonomic elements at scale and across different hierarchical levels. 

\noindent\textbf{High-Level Group Manager.} Low-level autonomic agents can interact through natural upstream and downstream dependencies or form a high-level service group managed by a high-level autonomic~\textbf{group manager}. This high-level group manager is often necessary for complex tasks requiring coordinated group actions, such as resolving complex service incidents in cloud environments. It is also easier and flexible to express goal-oriented terms in natural language at the high-level service group, such as ``\textit{reduce end-to-end service latency to 20 ms}''.

The LLM-based agent in the high-level group manager also uses the Plan-Execute feedback mechanism. Unlike the lower-level autonomic elements, the Planner in the high-level group manager breaks down complex management requests into sub-tasks and generates detailed, step-by-step plans. Each step corresponds to a specific low-level autonomic agent and includes executable code, such as assigning sub-tasks and analyzing collected metrics. The Executor then carries out these steps by executing the code. Feedback from the low-level autonomic agents is sent back to the high-level Planner to determine if the goals have been met. The Planner can adjust the plan based on this feedback and track the progress of the execution. This proactive approach allows the system to maintain optimal performance and reduce maintenance time, minimizing the need for direct human intervention.

\noindent \textbf{Three Working Mechanisms.}
\label{sec:working_mechanisms}
We define three working mechanisms with our hierarchical service management framework, each designed for particular task management and execution scenarios: a) Low-Level Autonomic Agent Working Alone; b) Multiple Low-Level Autonomic Agents Collaborating Under a Manager; c) Intra-Communication among Low-Level Autonomic Agents. To be more specific:

\noindent \textit{Low-Level Autonomic Agent Working Alone.} In this mode, the maintainer directly assign requests or tasks to a single low-level autonomic agent, requesting it to answer maintenance-related inquiries or take actions to fulfill maintenance requests. The agent operates independently to complete the assigned task. This mechanism is straightforward and effective for simple, isolated tasks that do not require coordination with other low-level autonomic agents.

\noindent \textit{Multiple Low-Level Autonomic Agents Collaborating Under a Manager.} In this mode, the high-level group manager either receives tasks from the maintainer or issues are raised by the low-level autonomic agents themselves. The manager first decomposes the task based on the received message, generate a plan, and then assigns the sub-tasks to the relevant low-level autonomic agents for execution. The execution results from each agent are collected, and the manager may modify or proceed to the next steps based on these outcomes. If the original plan cannot be followed, it is adjusted accordingly. This iterative process continues until the task is either completed or deemed unachievable. This mechanism is suitable for complex tasks that require collaborative efforts among multiple low-level autonomic agents.

\noindent \textit{Intra-Communication among Low-Level Autonomic Agents.} When a low-level autonomic agent encounters an issue it cannot resolve independently, it can seek assistance from other agents without involving of the high-level group manager. This mechanism facilitates internal communication among low-level autonomic agents, allowing them to collaborate and attempt to fix the issue by directly communicating with each other. 

To facilitate these working mechanisms, we use a message queue system as middleware for message passing and storage. This approach unifies the basic framework of all three working mechanisms, enhancing overall service availability and reliability. Additionally, the message queue system decouples message passing through asynchronous message queues, improving system robustness and flexibility. Note that we implement the first two mechanisms in the example below for simplicity and better illustration.

\noindent\textbf{Example Implementation on Sock Shop.} 
There are several microservce demos such as Online Boutique~\cite{onlineoutique} and Death Star~\cite{deathstar}. 
We select Sock Shop~\cite{sockshop} as the microservices demo application due to its comprehensive example of a real-world e-commerce application, modular microservices design, and Kubernetes-native setup. It offers a simple yet accessible start point to demonstrate sophisticated microservices and Kubernetes concepts. 

Sock Shop is a microservices demo application that emulates an e-commerce website for selling socks, designed to help users understand Kubernetes and microservices architecture. It consists of various microservices managed by Kubernetes, each handling different aspects of the e-commerce site, such as the website front-end, product catalog, orders, and payment (shown in Figure~\ref{fig:sock-shop-implementation} left). Each microservice is containerized using Docker, ensuring consistent and isolated deployment across different environments. These microservices are deployed and orchestrated by Kubernetes, which offers certain capabilities for managing them. However, Kubernetes relies on additional tools like Prometheus~\cite{prometheus_website} to perform tasks like monitoring and metrics collection. It also requires other essential declarative actions include auto-scaling, resource management, alerting, fault troubleshooting, and performance optimization. Currently, these actions are mostly performed manually by maintainers, requiring significant effort and domain expertise.

As shown in Figure~\ref{fig:sock-shop-implementation} right, we demonstrate the application of LLM-based autonomic management on the Sock Shop microservice demo project in Kubernetes. This autonomic management framework is based on the previously discussed LLM-based two-level hierarchical framework. Briefly, the high-level group manager oversees the entire group of microservices, interprets high-level tasks from maintainers, and breaks down these tasks into executable steps assigned to low-level autonomic agents. The high-level group manager also handles issues raised by low-level autonomic agents. Note that the high-level group manager does not directly operate any microservice; instead, each low-level autonomic agent manages its corresponding microservice, responding to tasks assigned by the high-level group manager, such as monitoring and maintenance. 

To be more specific, each Sock Shop containerized microservice is managed by a low-level LLM-based autonomic agent, making it an LLM-enhanced service component. Each agent is responsible for managing its own component. For example, the \texttt{Front-end} component is converted into the \texttt{Front-end} agent, enhanced by the LLM-based autonomic agent. Messages containing sub-tasks or maintainer requests are passed to the \texttt{Front-end} agent. Through the iterative process with the Planner and Executor, maintenance tasks are completed within the agent. Responses and feedback are then sent to the high-level group manager through the manager's message queue. If there is an unresolved problem, an issue is sent to the manager's message queue, seeking assistance from the high-level group manager (shown in the red line in Figure~\ref{fig:sock-shop-implementation}). The high-level group manager analyzes the message, makes necessary adjustments to the plan, and then transmits the revised steps to the message queues of the corresponding low-level autonomic agents, coordinating collaboration among various agents.

\section{Evaluation Benchmark}

After introducing the hierarchical LLM-based autonomic management framework and presenting an example ACV implementation, we now evaluate its performance to address the central question of this paper: \textbf{how close are we to achieving ACV with LLMs?} To answer this, we introduce a taxonomy categorizing tasks by their levels of autonomy and define specific task cases using Sock Shop.

\subsection{Autonomous Levels in Service Maintenance}
\label{sec:taxonomy}

\begin{figure*}[htb!]
  \includegraphics[width=0.85\textwidth]{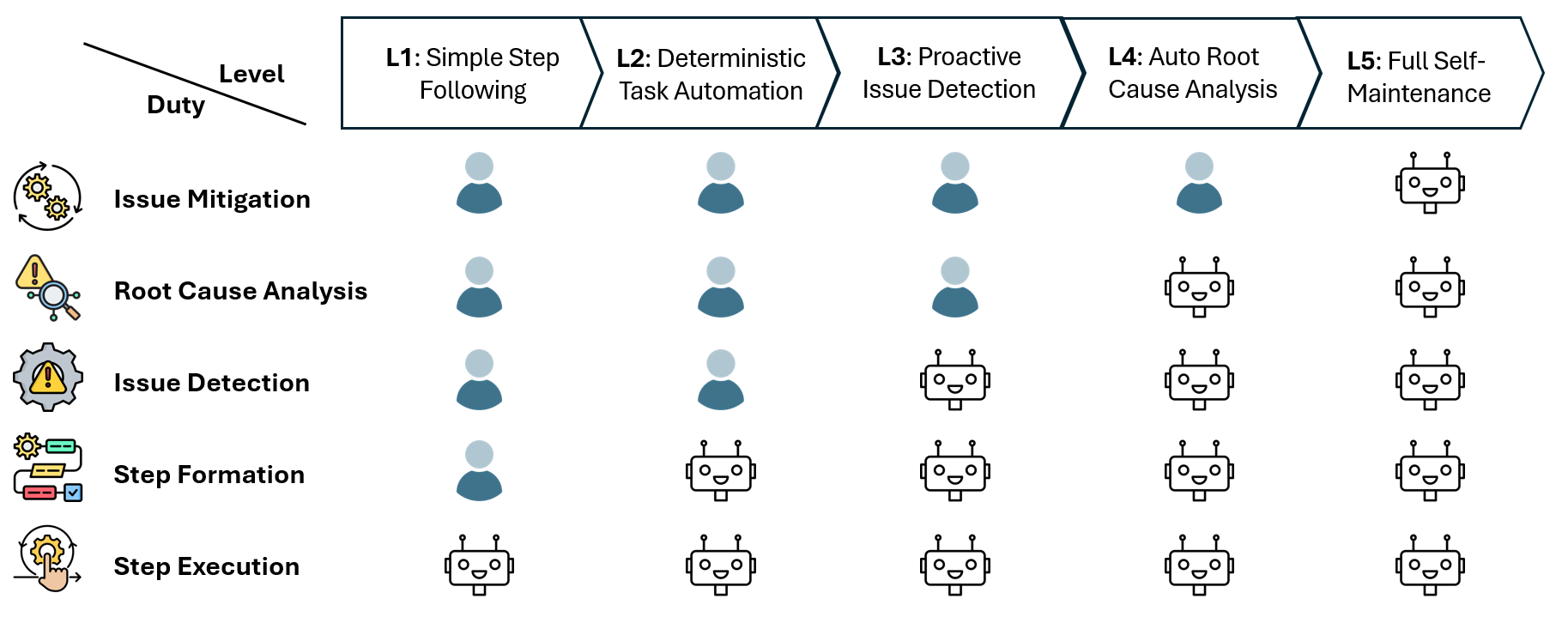}
  \caption{Taxonomy of autonomous levels in service maintenance, focusing on Self-Healing and Self-Optimization.}
  \label{fig:taxonomy}
\end{figure*}

Inspired by the six levels of autonomous driving and the categorization of Personal LLM Agents' duties based on intelligence levels in~\cite{Li2024PersonalLA}, we propose a taxonomy of five autonomy levels for service maintenance, depicted in Figure~\ref{fig:taxonomy}. 
The basic \textbf{L1} and \textbf{L2} levels represent the foundational maintenance capabilities of autonomic management agents, such as understanding users' intent and possessing necessary service maintenance knowledge (e.g., writing correct maintenance code using \texttt{kubectl} commands). At Level 1 (``Simple Step Following''), we assess whether the agent can determine correct operational commands to fulfill specific imperative instructions, such as scaling replicas to three. Level 2 (``Deterministic Task Automation'') additionally requires the agent to possess the planning capability by decomposing a complex task into smaller executable steps, such as checking the service health status by generating a series of metric collection and understanding steps.

L1 and L2 levels are characterized by imperative tasks, focusing on fulfilling specified tasks. Higher autonomy levels require the agent to fulfill declarative tasks, proactively performing actions to achieve predefined goals or states. These high-level goals align with the four self-management objectives outlined in the ACV paper. For simplicity, this work focus on tasks related to Self-Optimization and Self-Healing, leaving the other two areas for future research.

Achieving Self-Optimization and Self-Healing necessitates three key capabilities in the autonomic management agent: ``Proactive Issue Detection'', ``Automatic Root cause Analysis'', and ``Full Self-Maintenance'' with the generation and execution of mitigation solutions. These capabilities correspond to \textbf{L3}, \textbf{L4}, and \textbf{L5} levels in our taxonomy shown in Figure~\ref{fig:taxonomy}. With this five-level taxonomy established, we proceed to design specific evaluation tasks tailored to each level within the Sock Shop context. These tasks will serve as benchmarks to quantitatively assess the capabilities of current LLMs in advancing towards the realization of ACV. 

\subsection{Online Live Evaluation Benchmark}
\label{sec:task_def}

\begin{table*}[t]
  \centering
  \small
  \begin{tabular}{c | c | c | c | c | c}
  \hline
  \hline
  Request & Operation & \multirow{2}{*}{Level} & \multirow{2}{*}{Task Name} & \multirow{2}{*}{Task Description} & \multirow{2}{*}{Traffic Load} \\
  Level & Type & & & & \\
  \hline
  \multirow{14}{*}{Low} & DCM & L1 & Deployment Creation & Create a deployment using \texttt{Catalogue}'s deployment \texttt{YAML} file. & Moderate on \texttt{Cat.}\\
   & DCM & L1 & Deployment Update & Update \texttt{Catalogue}'s environment variable \texttt{logger\_flag} to true. & Moderate on \texttt{Cat.}\\
   & DCM & L1 & Roll Back & Rollback \texttt{Catalogue}'s image version to \texttt{0.3.4}. & Moderate on \texttt{Cat.}\\
  & DCM & L1 & Pod Restart & Restart all the pods of \texttt{Catalogue} immediately. & Moderate on \texttt{Cat.}\\
  & RM & L1 & Metric Collection-CPU & Report the CPU usage of \texttt{Catalogue}. & Moderate on \texttt{Cat.}\\
  & RM & L1 & Metric Collection-Latency & Report the P99 latency of \texttt{Catalogue}. & Moderate on \texttt{Cat.}\\
  & RM & L1 & Manual Scaling & Scale \texttt{Catalogue}'s replicas to 3. & Heavy on \texttt{Cat.}\\
  \cline{2-6}
  & RM & L2 & Health Check & Check \texttt{Catalogue}'s healthy status immediately. & Heavy on \texttt{Cat.}\\
  & RM & L2 & Performance Check & Check whether \texttt{Catalogue}'s performance is normal immediately. & Moderate on \texttt{Cat.}\\
  & RM & L2 & Auto Scaling & Implement auto-scaling with sensible thresholds for \texttt{Catalogue}. & Heavy on \texttt{Cat.}\\
 & RM & L2 & Latency Reduction & Reduce the P99 latency of \texttt{Catalogue} to under 300 ms. & Heavy on \texttt{Cat.}\\
  & RM & L2 & CPU Reduction & \makecell{Reduce the average CPU usage of \texttt{Catalogue} to \\ under $30\%$ of resource limit.} & Moderate on \texttt{Cat.}\\
  \hline
  % high
  \multirow{7}{*}{High} & RM & L2 & Latency Reduction-Cat. & Reduce the P99 latency of \texttt{Catalogue} to under 300 ms. & Heavy on \texttt{FE.}\\
  & RM & L2 & CPU Reduction-Cat. & \makecell{Reduce the average CPU usage of \texttt{Catalogue} to \\ under $30\%$ of resource limit.} & Heavy on \texttt{FE.}\\
  & RM & L2 & Latency Reduction-Group & \makecell{Reduce the total P99 latency of \texttt{Catalogue} \\ and \texttt{Front-end} to under 400 ms.} & Heavy on \texttt{FE.}\\
  & RM & L2 & CPU Reduction-Group & \makecell{Reduce each \texttt{Catalogue} and \\ \texttt{Front-end}'s average CPU usage to under $30\%$.} & Heavy on \texttt{FE.}\\  
  \hline
  \hline
  \multicolumn{6}{l}{\emph{Note1:} No initial deployment of \texttt{Catalogue} for the Deployment Creation task, and for the Roll Back task, the initial pod image version is \texttt{0.3.6}. } \\
  \multicolumn{6}{l}{\emph{Note2:} Definition of each Traffic Load can be found in Table~\ref{tb:traffic}. \texttt{Cat} and \texttt{FE} are abbreviations for \texttt{Catalogue} and \texttt{Front-end}, respectively.} 
  \end{tabular}
  \caption{Definition of L1 and L2 tasks within the context of Sock Shop.}
  \label{tb:L1_L2_def}
  \vspace{-4mm}
\end{table*}

In traditional benchmarking, evaluations are typically performed against a specific dataset in an offline manner. However, to assess whether an agent can achieve the complete Detection-RCA-Mitigation maintenance cycle, we need an \textbf{Online Live Evaluation Benchmark} that operates within a functional service environment. To create this environment, we deploy the Sock Shop service using Kubernetes and simulate traffic to ensure it functions as a live system. This setup allows us to introduce various tasks, such as metric collection and health checks, to evaluate the basic L1 and L2 capabilities. For evaluating L3, L4, and L5 capabilities, we further employ \textbf{Chaos Engineering} techniques~\cite{principles_of_chaos} to intentionally inject faults or induce performance issues.

Table~\ref{tb:L1_L2_def} lists 16 evaluation tasks for L1 and L2 levels, distinguishing between 12 tasks for low-level and 4 tasks for high-level group manager. Low-level tasks are directly applied to agents that manage individual service component, while high-level tasks involve the manager agent, allowing us to examine individual and collaborative task handling. For this study, we focus on the \texttt{Catalogue} for low-level evaluation and include the high-level group manager and \texttt{Front-end} for high-level evaluation tasks. These tasks encompass basic Deployment and Creation Management (DCM) operations as well as Runtime Management (RM) activities, reflecting common microservice management operations. Traffic load levels are specified to ensure tasks, such as reducing latency, are meaningful.

In contrast, L3, L4, and L5 evaluations involve injecting specific faults or issues to trigger the evaluation. We introduce three types of faults and performance issues:
\begin{itemize}[noitemsep, left=0pt]
    \item \textbf{Pod Failure}: Replace \texttt{Catalogue} pod image with a fake and non-functional one. 
    \item \textbf{CPU Stress}: Occupy 100\% of \texttt{Catalogue} pod's CPU resources. 
    \item \textbf{Rising Traffic}: Gradually increase traffic, leading to high resource usage and extended service latency. This pattern is applied directly to \texttt{Catalogue} for low-level tasks and to \texttt{Front-end} for high-level tasks. 
\end{itemize}
While many other faults and issues exist in microservice management, these three are representative and pertinent for this study.

We then evaluate the system's ability to perform self-management operations to meet predefined Service Level Objectives (SLOs) under these injected conditions. Within the Sock Shop context, we define the following SLOs:
\begin{enumerate}[noitemsep, left=0pt]
    \item All service components maintain a healthy READY state.
    \item CPU and memory usage for each component remains under 50\% of allocated resources.;
    \item The P99 latency for each component is stable, with an average P99 latency below 200ms.
\end{enumerate}
Evaluation of L3/L4/L5 tasks are also distinguished by both low-level autonomic agents and high-level group manager, with the above SLOs communicated to these agents based on task requirements. Given that L3, L4, and L5 tasks are typically executed in a unified Detection-RCA-Mitigation sequence, we implement a combined task to evaluate all three levels simultaneously. A detailed description of these tasks is provided in Section~\ref{sec:task_description_L3_4_5} in the Appendix.

\subsection{Evaluation Metric}
\label{sec:evaluation_metric}

Our evaluation focuses on two main types of metrics: efficiency and quality. For efficiency metrics, we monitor the number of steps taken in each evaluation, with each step defined as a single call to the base LLM. We also track the number of communication rounds between high-level group manager and low-level autonomic agents for tasks initiated by the high-level group manager. Steps that result in execution errors are included as well, as these errors reflect the base LLM's ability to generate correct service maintenance code.

On the quality side, we measure the task fulfillment rate at various levels. For L1 and L2 tasks, we assess whether the assigned tasks are completed successfully. For L3 tasks, we evaluate the agent's ability to correctly judge if the system meets the SLOs. In L4 tasks, we determine if the management agents can identify the correct root causes. Finally, for L5 tasks, we check whether injected faults or issues are successfully mitigated. Additionally, for tasks assigned to high-level group manager, we evaluate the high-level agent's ability to correctly delegate tasks to low-level autonomic agents.

\section{Experiment}

\subsection{Setup}

\noindent\textbf{Sock Shop Deployment.}  Sock Shop is deployed on a Minikube~\cite{minikube} cluster with 6 cores and 16 GB of memory, utilizing an Intel Xeon Gold 6338 CPU @ 2 GHz server. The deployment includes an enabled metric server and a Prometheus server, configured according to the Sock Shop deployment files. Load simulation is applied using Locust~\cite{locust}, with traffic levels detailed in Table~\ref{tb:traffic} in the Appendix. The faults of Pod Failure and CPU Stress are injected by utilizing the Chaos Mesh tool~\cite{chaos_mesh}. 

\noindent\textbf{Agent-based Management System.} The agent-based management system introduced in Section~\ref{sec:framework} is implemented by leveraging the multi-agent framework AutoGen~\cite{wu2023autogen}. It incorporates agent groups consisting of LLM-based agents and non-LLM-based code executors for each microservice component, serving as low-level autonomic agents. Similarly, the high-level group manager comprises a LLM-based agent and a non-LLM-based code executor. Communication among these agents is facilitated by RabbitMQ \cite{rabbitmq}, and GPT-4 Turbo is adopted as the underlying LLM.

The prompts for low-level autonomic agents and high-level group manager are given in Sections~\ref{sec:prompt} in the Appendix. Most of instructions in them are generally applicable to other microservices managed by Kubernetes. However, in the prompt for low-level autonomic agents, we include instructions that is tailored specifically to the Sock Shop service. Those instructions are effective at eliminating inefficient self-correcting steps. For example, agents might use the wrong label selector \texttt{``-l app=Catalogue''} instead of the correct \texttt{``-l name=Catalogue''} to identify \texttt{Catalogue}. While agents can correct such errors through several self-correction rounds, these mistakes significantly disrupt the execution flow since many \texttt{kubectl} commands depend on accurate label selectors. Similarly, querying Prometheus metrics correctly is often challenging because it requires precise input of metric names and filters, which are service-specific and not typically known to LLMs. Although these specific instructions reduce generality, they are necessary, especially for low-level autonomic agents managing elements with domain-specific information that LLMs lack.

\noindent\textbf{Experiment Procedure.} As detailed in Section~\ref{sec:task_def}, tasks are categorized for management agents at both low and high levels. Given the distinct characteristics of L1/L2 and L3/L4/L5 tasks, we establish four distinct experiment configurations, each outlined comprehensively in Table~\ref{tb:exp_setting}. Overall, the procedure of each experiment run involves the environment setup (i.e., deploying Sock Shop and ensuring stable traffic), sending task requirement to corresponding management agents and finally evaluating the task performance. All experiments are randomly repeated three times to reduce statistical fluctuation. For L3/L4/L5 tasks managed by low-level autonomic agents, three consecutive evaluations are allowed to account for potential missed detections in initial trials. Conversely, for L3/L4/L5 tasks managed by high-level agents, a single evaluation suffices as multiple agents collaborate to detect issues.

\begin{table}[t]
  \centering
  \small
  \begin{tabular}{c | c | c }
  \hline
  \hline
  & Low-Level & High-Level \\
  \hline
  \makecell{L1\\L2} & \makecell[l]{1. Deploy Sock Shop and \\ensure stable traffic; \\ 2. Send task to \texttt{Catalogue}'s \\ management agent; \\ 3. Task evaluation; \\ 4. Repeat the above 3 times.} & \makecell[l]{1. Deploy Sock Shop and \\ensure stable traffic;\\ 2. Send task to high-level \\ manager; \\ 3. Task evaluation; \\ 4. Repeat the above 3 times.} \\
\hline
 \makecell{L3\\L4\\L5} &  \makecell[l]{1. Deploy Sock Shop and \\ensure stable traffic; \\ 2. Inject faults/issues; \\ 3. Send task to \texttt{Catalogue}'s \\ management agent; \\ 4. Perform 3 consecutive evals; \\ 5. Repeat the above 3 times.} & \makecell[l]{1. Deploy Sock Shop and \\ensure stable traffic; \\ 2. Inject faults/issues; \\ 3. Send task to high-level \\ manager; \\ 4. Perform one evaluation; \\ 5. Repeat the above 3 times.} \\
  \hline
  \hline
  \end{tabular}
  \caption{Experiment procedures for different configurations.}
  \label{tb:exp_setting}
\end{table}

\noindent\textbf{Evaluation}. The evaluation metrics discussed in Section~\ref{sec:evaluation_metric} were gathered through a manual review of the agents' chat history. This review was performed independently by two authors with expertise in Kubernetes service maintenance. A sample log of the agents' chat history corresponding to the Latency Reduction task applied to \texttt{Catalogue} can be accessed via this link.\footnote{https://paste.ubuntu.com/p/HnnGJ9qPgM/}

\subsection{Results for Tasks Applied to the Low-Level Autonomic Agent}
 
\begin{table}[t]
  \centering
  \small
  \begin{tabular}{c | c | c  c  c }
  \hline
  \hline
  & Task Name & Is Passed? & Steps & \makecell{Steps with \\Exec. Error} \\
  \hline
  \multirow{7}{*}{L1} & Deployment Creation & [1,1,1] & [3,4,4] & [0,0,0]  \\
  & Deployment Update & [1,1,1] & [5,5,5] & [0,0,0]  \\
  & Roll back & [1,1,1] & [5,9,6] & [0,0,0]  \\
  & Pod Restart & [1,1,1] & [4,5,5] & [0,0,0]  \\
  & Metric Collection-CPU & [1,1,1] & [5,5,4] & [0,0,0]  \\
  & Metric Collection-Latency & [1,1,1] & [3,3,3] & [0,0,0]  \\
  & Manual Scaling & [1,1,1] & [9,5,6] & [0,0,0]  \\
  \hline
  \multirow{5}{*}{L2}  & Health Check & [1,1,1] & [7,7,4] & [0,0,0]  \\
  & Performance Check & [1,1,1] & [4,4,4] & [1,0,0] \\
  & Auto Scaling & [1,1,1] & [6,11,4] & [0,2,0] \\
  & Latency Reduction & [1,1,0] & [11,13,15] & [0,0,1] \\
  & CPU Reduction & [1,0,1] & [11,10,8] & [0,0,0] \\
  \hline
\textbf{L1}  & \textsc{Average} & \textbf{1.0} & \textbf{4.9} & \textbf{0} \\
\textbf{L2}  & \textsc{Average} & \textbf{0.87} & \textbf{7.9} & \textbf{0.3} \\
  \hline
  \hline
  \end{tabular}
  \caption{Results for L1 and L2 tasks applied to the low-level autonomic agent of \texttt{Catalogue}. Each element in the 3-tuple list corresponds to one of the 3 repeated experiment runs.}
  \label{tb:ind_L1_L2_res}
  \vspace{-4mm}
\end{table}

Table~\ref{tb:ind_L1_L2_res} presents the experimental results for the L1 and L2 tasks applied on the low-level autonomic agent of \texttt{Catalogue}. Details include the pass status, the number of steps taken, and any steps with execution errors. The key findings are summarized as follows:

\begin{itemize}[noitemsep, left=0pt]
\item The L1 and L2 tasks demonstrated high task completion rates, achieving 100\% for L1 and 87\% for L2. One of the failed L2 experiments was the Latency Reduction, which failed because an ineffective action was made by editing another service configuration. The other unsuccessful experiment, CPU Reduction, failed due to misguided actions, particularly the reduction of CPU requests and limits, resulting in non-operational pods.
\item On average, the L1 task required approximately 5 steps to complete, while the L2 task required 8 steps, indicating that L2 tasks often involve additional planning steps. While core operations often required only 1 or 2 steps, those extra steps were related to precautionary checks before and after taking actions.
\item Code execution errors were minimal, and the system generally self-corrected these errors, underscoring the robustness of the LLM-based management system.
\end{itemize}

These results indicate that low-level LLM-based autonomic agents are highly effective in performing basic service maintenance tasks in Kubernetes, establishing a solid foundation for advancing towards higher levels of autonomy.

\begin{table}[t]
  \centering
  \small
  \begin{tabular}{c | c | c  c  c c c c}
  \hline
  \hline
  \multirow{2}{*}{\makecell{Injected\\Faults}} & \multirow{2}{*}{\makecell{Trial \\Index}} & \multirow{2}{*}{\makecell{Steps\\(3 Evals)}} & \multicolumn{5}{c}{Is Task Passed?} \\
  \cline{4-8}
  & & & \makecell{L3} & \makecell{L4} & \makecell{L5} & \makecell{First} & \makecell{Overall} \\
  \hline
  \multirow{3}{*}{\makecell{Pod\\Failure}} & 1 & [7,17,15] & 3/3 & 2/3 & 0/3 & 0 & 0\\
  & 2 & [9,11,11] & 2/3 & 1/1 & 1/1 & 1 & 1\\
  & 3 & [21,6,7] & 3/3 & 1/1 & 1/1 & 1 & 1\\
  \hline
  \multirow{3}{*}{\makecell{CPU\\Stress}} & 1 & [7,9,9] & 2/3 & 1/2 & 1/2 & 0 & 1\\
  & 2 & [13,11,12] & 3/3 & 0/3 & 0/3 & 0 & 0\\
  & 3 & [4,9,1] & 1/3 & 0/1 & 0/1 & 0 & 0\\
  \hline
  \multirow{3}{*}{\makecell{Rising\\Traffic}} & 1 & [4,4,1] & 2/3 & 0/0 & 0/0 & 0 & 0\\
  & 2 & [18,7,7] & 3/3 & 0/0 & 0/0 & - & -\\
  & 3 & [13,7,4] & 1/3 & 1/1 & 0/1 & 0 & 0\\
  \hline
 \multicolumn{2}{c|}{\textsc{Average}} & \textbf{8.8} (per eval) & \textbf{0.74} & \textbf{0.5} & \textbf{0.25} & \textbf{0.25} & \textbf{0.38}\\
\hline
\hline
  \end{tabular}
  \caption{Result summary for L3/L4/L5 tasks applied to the low-level autonomic agent of \texttt{Catalogue}. ``Trial Index'' represents each experiment run, and the element in the 3-tuple list of ``Steps'' corresponds to one of the three consecutive evaluations. The ``x/y'' pattern in the ``Is Task Passed?'' columns denotes that ``x'' out of ``y'' cases are passed. More result details can be found in Table~\ref{tb:ind_L3_4_5}.}
  \label{tb:ind_L3_4_5_summary}
  \vspace{-3mm}
\end{table}

The results for L3, L4, and L5 tasks applied to the low-level autonomic agent of \texttt{Catalogue} are summarized in Table~~\ref{tb:ind_L3_4_5_summary}. Detailed evaluations, including failure reasons, are provided in Table~~\ref{tb:ind_L3_4_5} in the Appendix. Table~\ref{tb:ind_L3_4_5_summary} presents the number of steps for three consecutive evaluations per experiment run and assesses task completion at different granularities: detection of failures/issues (L3), identification of root causes (L4), issue mitigation (L5), and task resolution in the first evaluation (First) or overall after three evaluations (Overall). Notably, not all evaluations involved valid L4/L5 tasks, as some issues were resolved prior to evaluation due to previous interventions or unsuccessful fault injection (e.g., Trial 2 in the Rising Traffic case). Key observations from Table~\ref{tb:ind_L3_4_5_summary} include:
\begin{itemize}[noitemsep, left=0pt]
    \item On average, approximately 9 steps were needed to perform L3/L4/L5 tasks, indicating higher complexity compared to L1/L2 tasks. The number of required steps varied widely across the three evaluations in one experiment run. This variation is attributed to the necessity of the full Detection-RCA-Mitigation cycle (e.g., some evaluation terminated early if no issue was detected).
    \item The task pass rates decreased from 0.74 (L3) to 0.5 (L4) and 0.25 (L5), reflecting the increased complexity of higher-level self-management tasks. Coupled with the high completion rates for L1 and L2 tasks, we conclude that \textbf{the low-level autonomic agent can achieve Level 3 autonomy in Self-Optimization and Self-Healing}, according to the taxonomy in Section~\ref{sec:taxonomy}. Although full L5 autonomy has not been achieved by current LLMs, these results demonstrate significant potential for LLMs in advancing ACV.
    \item The task completion rate improved from 0.25 to 0.38 when evaluations were allowed to be performed up to three times. This indicates that repeated Detection-RCA-Mitigation cycles enhance the system's ability to mitigate issues.
\end{itemize}

\subsection{Results for Tasks Applied to the High-Level Group Manager}

\begin{table}[t]
  \centering
  \small
  \begin{tabular}{c | c | c  c  c c }
  \hline
  \hline
  \makecell{Task\\Name} & \makecell{Trial\\Index} & \makecell{Rounds} & \makecell{Steps} & \makecell{Is Correct Task\\Assignment?} & \makecell{Is Task\\Passed?}\\
  \hline
  \multirow{3}{*}{\makecell{Latency \\Reduction\\-Cat.}} & 1 & 3 & 19 & 3/3 & 1 \\
  & 2 & 4 & 35 & 4/4 & 1 \\
  & 3 & 2 & 8 & 2/2 & 1\\
  \hline
  \multirow{3}{*}{\makecell{CPU \\Reduction\\-Cat.} } & 1 & 3 & 19 & 2/3 & 0 \\
  & 2 & 3 & 27 & 3/3 & 1 \\
  & $3$ & 3 & 19 & 2/3 & 0\\
  \hline
  \multirow{3}{*}{\makecell{Latency \\Reduction\\-Group}} & 1 & 3 & 29 & 3/3 & 1 \\
  & 2 & 3 & 46 & 3/3 & 1 \\
  & 3 & 3 & 36 & 2/3 & 1\\
  \hline
  \multirow{3}{*}{\makecell{CPU \\Reduction\\-Group}} & 1 & 3 & 35 & 1/3 & 0 \\
  & 2 & 3 & 36 & 3/3 & 0 \\
  & 3 & 5 & 63 & 5/5 & 0\\
  \hline
 \multicolumn{2}{c|}{\textsc{Average}} & \textbf{3.2} & \textbf{31} & \textbf{0.87} & \textbf{0.58}\\
\hline
\hline
  \end{tabular}
  \caption{Result summary for L2 tasks applied to the high-level group manager. More details are given in Table~\ref{tb:group_L1_2}.}
  \label{tb:group_L1_L2_summary}
\end{table}

\begin{figure*}[htb!]
 \includegraphics[width=0.9\textwidth]{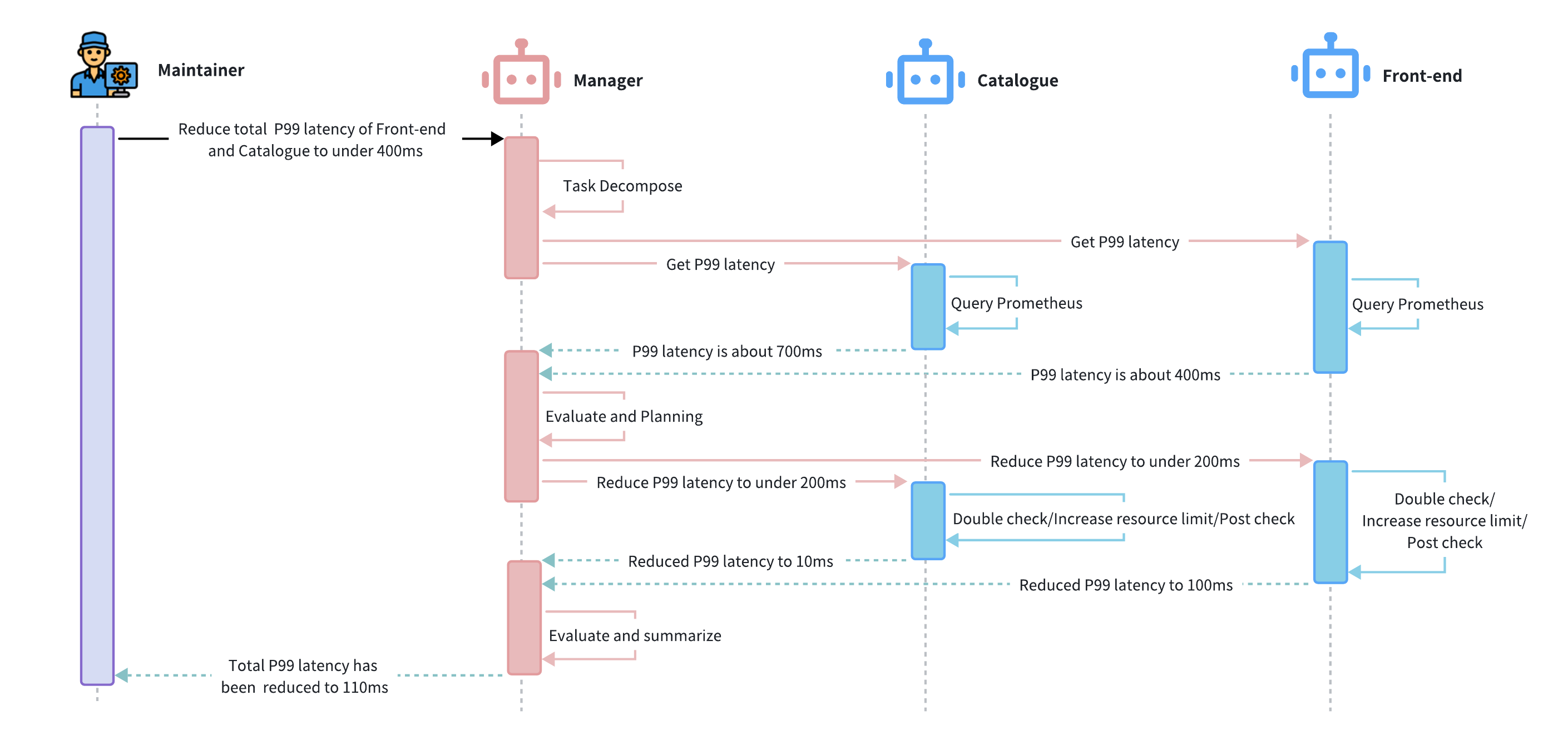}
  \caption {Sequence diagram for the Latency Reduction-Group task applied to the high-level group manager.}
  \label{fig:latencytask}
  % \vspace{-3mm}
\end{figure*}

We next present the results of applying tasks to the high-level group manager, focusing on the task completion rate, task assignment to low-level autonomic agents, and how these agents fulfill sub-tasks.

Table~\ref{tb:group_L1_L2_summary}  summarizes the results for L2 tasks applied to the high-level group manager, with detailed results in Table~\ref{tb:group_L1_2}. Besides the number of steps and the overall task pass status, Table~\ref{tb:group_L1_L2_summary} also indicates the number of communication rounds between high-level and low-level autonomic agents and the accuracy of task assignment by the high-level agent. Key observations include:
\begin{itemize}[noitemsep, left=0pt]
\item On average, it took about 3 rounds to complete each L2 task. These rounds typically involve: i) the first round of collecting metrics to determine task necessity; ii) the second round where low-level autonomic agents perform the actual task; iii) the final round of summarizing and reporting results, and terminating the task. For instance, Figure~\ref{fig:latencytask} illustrates the sequence for the Latency Reduction-Group task. Upon receiving the task to reduce P99 latency of \texttt{Front-end} and \texttt{Catalogue} to under 400 ms, the high-level group manager first gathered P99 latency metrics for both services. After finding that the latencies were about 400 ms and 700 ms respectively, the manager assigned the task of reducing latency to under 200 ms. Low-level autonomic agents carry out the required actions and reported back, enabling the high-level group manager to summarize and terminate the task.
\item Each task took approximately 31 steps to complete, significantly higher than the 8 steps required for L2 tasks handled by low-level autonomic agents (Table~\ref{tb:ind_L1_L2_res}). The increase in steps is due to the higher complexity of tasks involving multiple low-level autonomic agents (e.g., Latency Reduction-Group task) and additional planning and communication steps required in the hierarchical management architecture. Notably, the steps taken by low-level autonomic agents during task execution (e.g., the round for the latency reduction task) were comparable to those in Table~\ref{tb:ind_L1_L2_res}. Thus, since a hierarchical management structure introduces extra planning and communication costs, directly assigning maintenance requests to specific low-level autonomic agents might sometimes be more efficient.
\item The high-level management system demonstrates strong task assignment capabilities, achieving an accuracy of 0.87 for L2 tasks. Analysis of task completion reveals that while latency reduction tasks exhibit high success rates, tasks aimed at CPU reduction suffer substantial accuracy declines. Detailed examination of failure instances, detailed in Table~\ref{tb:group_L1_2}, highlights common issues such as erroneous CPU request reductions (mirroring the L2 Latency Reduction task failures in Table~\ref{tb:ind_L1_L2_res}) and horizontal scaling actions insufficient for task fulfillment.
\end{itemize}

\begin{table}[t]
  \centering
  \small
  \begin{tabular}{c | c | c  c c  c c c }
  \hline
  \hline
  \multirow{2}{*}{\makecell{Injected\\Faults}} & \multirow{2}{*}{\makecell{Trial \\Index}} & \multirow{2}{*}{Rounds} & \multirow{2}{*}{\makecell{Steps}} & \multicolumn{4}{c}{Is Task Passed?} \\
  \cline{5-8}
  & & & & \makecell{L3} & \makecell{L4} & \makecell{L5} & Overall\\
  \hline
  \multirow{3}{*}{\makecell{Pod\\Failure}} & 1 & 2 & 28 & 1/1 & 1/1 & 1/1 & 1 \\
  & 2 & 3 & 31 & 2/2 & 0/1 & 1/1 & 1 \\
  & 3 & 2 & 30 & 1/1 & 0/1 & 1/1 & 1 \\
  \hline
  \multirow{3}{*}{\makecell{CPU\\Stress}} & 1 & 5 & 61 & 3/4 & 1/3 & 1/3 & 1 \\
  & 2 & 4 & 69 & 0/0 & 0/0 & 2/4 & 1 \\
  & 3 & 4 & 50 & 0/0 & 0/0 & 2/3 & 1 \\
  \hline
  \multirow{3}{*}{\makecell{Rising\\Traffic}} & 1 & 4 & 66 & 0/0 & 0/0 & 0/4 & 0 \\
  & 2 & 2 & 21 & 0/0 & 0/0 & 0/0 & 0 \\
  & 3 & 4 & 49 & 2/2 & 2/2 & 0/2 & 0 \\
  \hline
 \multicolumn{2}{c|}{\textsc{Average}} &  \textbf{3.3} & \textbf{45} & \textbf{0.9} & \textbf{0.5} & \textbf{0.42} & \textbf{0.67}\\
\hline
\hline
  \end{tabular}
  \caption{Result summary for L3/L4/L5 tasks applied to the high-level group manager. The ``x/y'' pattern in the ``Is Task Passed?'' columns denotes that ``x'' out of ``y'' cases are passed. More result details can be found in Table~\ref{tb:group_L3_4_5}.}
  \label{tb:group_L3_4_5_summary}
  \vspace{-4mm}
\end{table}

The results for the L3, L4, and L5 tasks are summarized in Table~\ref{tb:group_L3_4_5_summary}, with detailed information provided in Table~~\ref{tb:group_L3_4_5}. These tables list the number of rounds and steps performed, as well as the task completion rates at each level and overall, including whether the injected faults were resolved. Key findings include:
\begin{itemize}[noitemsep, left=0pt]
    \item The average number of rounds required was approximately 3, similar to the L2 task. The procedure for these rounds typically followed the similar pattern of metric collection, task execution, and result summary. However, additional rounds of task execution were sometimes necessary if the previous ones did not complete the task.
    \item On average, 45 steps were required, significantly more than the 31 steps needed for L2 tasks. This increase may be due to the greater complexity and deeper analysis required for each round of the L3/L4/L5 sub-tasks.
    \item The overall task completion rate decreased from 0.9 (L3) to 0.5 (L4) and 0.42 (L5), consistent with the pattern observed in low-level autonomic agent tasks. However, the actual completion rates were higher than those for the corresponding low-level tasks, suggesting that including a hierarchical management structure may improve the overall task completion rate. This observation further confirms our previous proposition that attaining Level 3 autonomy is feasible with the existing LLM-based agent management system.
\end{itemize}

\subsection{Failure Analysis}

\begin{table*}[t]
  \centering
  \small
  \begin{tabular}{c | l | c | c}
  \hline
  \hline
  \makecell{Task with Failure} & \makecell[c]{Failure Reason} & \makecell{Failure\\Count} & Potential Solution Direction\\
  \hline
  \multirow{3}{*}{\makecell{L3\\(Wrong Detection)}} & hallucination during numerical comparison & 4 & \makecell{reduce hallucination} \\
  \cline{2-4}
  & task omission & 3 & better instruction following \\
  \cline{2-4}
  & \makecell[l]{mis-guided by previous action execution output} & 1 & improve reasoning \\
  \hline
  \multirow{3}{*}{\makecell{L4\\(Wrong RCA)}} & \makecell[l]{unable to collect important information (e.g., CPU limit)} & 5 & \makecell{better reasoning; more domain knowledge} \\
  \cline{2-4}
  & \makecell[l]{mis-guided by previous action execution output} & 3 & improve reasoning \\
  \cline{2-4}
  & task omission & 2 & better instruction following \\
  \hline
  \multirow{5}{*}{\makecell{L5\\(Failed Mitigation)}} & ineffective action & 4 & \makecell{improve reasoning}\\
  \cline{2-4}
  & adopted action insufficient to resolve issue & 4 & \makecell{improve reasoning} \\
  \cline{2-4}
  & \makecell[l]{unable to report correctly (e.g., hallucinated report, omit report)} & 4 & \makecell{better instruction following; reduce hallucination} \\  
  \cline{2-4}
  & wrong choice of action & 3 & \makecell{improve reasoning} \\
  \cline{2-4}
  & \makecell[l]{failed to execute action (e.g., action generated but not applied)} & 3 & \makecell{better instruction following} \\
  \hline
  \multirow{3}{*}{\makecell{Wrong High-level\\Task Assignment}} & task omission & 7 & better instruction following\\
  \cline{2-4}
  & affected by incorrect low-level report & 5 & improve reporting at low-level\\
  \cline{2-4}
  & wrong reasoning based on low-level report & 2 & improve reasoning\\
\hline
\hline
  \end{tabular}
  \caption{Failure reason analysis and potential solution directions for tasks with failures.}
  \label{tb:error_analysis}
  \vspace{-1.5em}

\end{table*}

Lastly, we present an in-depth analysis of failed cases in tasks related to L3 (incorrect detection), L4 (incorrect root cause analysis), L5 (mitigation failure), and erroneous high-level task assignment. The detailed results, categorized by failure type and potential improvement directions, are shown in Table~\ref{tb:error_analysis}. These failures stem from issues such as instruction-following errors (e.g., task omission), hallucinations, deficiencies in reasoning capabilities, and insufficient domain knowledge.

To address these issues, we propose two primary strategies. First, we can employ a superior base LLM, either a state-of-the-art general LLM or a domain-specific LLM fine-tuned with service maintenance knowledge. Second, we can enhance our existing agent framework by integrating additional modules. These modules might include mechanisms for incorporating domain knowledge using retrieval-augmented generation techniques~\cite{NEURIPS2020_6b493230} or the addition of critic agents to mitigate hallucinations and bolster reasoning capabilities.

\section{Discussion}

This study systematically evaluates the feasibility of realizing autonomic computing using LLM-based agents. Our comprehensive experiments, conducted with the microservice Sock Shop in Kubernetes, highlight several key topics that merit further discussion.

\noindent\textbf{Alternative LLM-based Management System with Agents.} We utilize a hierarchical multi-agent system to decompose tasks at different levels, employing the Plan-Execution feedback mechanism in each agent for autonomic task management. Other agent-based solutions could also be considered, such as integrating critic agents to oversee the entire process, implementing a memory mechanism to store short/long-term maintenance history, and including a lifelong self-learning module to continuously enhance agent performance~\cite{qiao2023taskweaver, wu2023autogen, Wang2023Voyager}. Additionally, different LLMs could be adopted as base models. While GPT-4 Turbo serves as a capable representative, other LLMs might exhibit different behaviors in service maintenance contexts. Fine-tuning LLMs on service maintenance-related data is also an existing area worth exploration. 

\noindent\textbf{Evaluation Benchmark Enrichment.} The benchmark tasks in this study focus on Self-Optimization and Self-Healing. While this is a first attempt to systematically assess ACV using LLM-based agents, expanding the scope of the evaluation benchmark in future work would be valuable. Including Self-Configuration and Self-Protection would demonstrate the system's flexibility, evolution capability, and resilience in hostile cyber environments. Additionally, while we aim to cover most failure or issue types, some, such as grey failures~\cite{Huang2017grey} and metastable failures~\cite{Huang2022meta}, are not included. Further, although the Sock Shop microservice covers major aspects of microservice management using Kubernetes, it lacks the complexity, scalability, and diversity of real-world large-scale cloud services. Evaluating more diverse benchmarks, such as Online Boutique~\cite{onlineoutique} and Death Star~\cite{deathstar}, and those beyond microservice management, would be a logical next step. Finally, although Section~\ref{sec:framework} outlines the intra-communication mechanisms among low-level autonomic agents, concrete evaluation tasks are needed to assess agents' collaborative issue-solving capabilities in self-organizing scenarios without high-level supervision. Such patterns are crucial for managing extensive distributed computing systems that do not have centralized management.

\noindent\textbf{Application in Practice.} Two immediate real-world implications of our research are discernible. Firstly, our agent-based management system, initially developed for Sock Shop, could be generalized as a standard component for existing Kubernetes management frameworks. This abstraction could manifest as middleware tools akin to Prometheus and other metric/log collection tools, seamlessly transforming each microservice component into autonomic units during application development. Such integration would augment Kubernetes' capabilities in declarative, autonomic, and proactive management of service deployments. Second, building a live evaluation benchmark would be an effective way to facilitate replicating real-world issues and assessing management systems' ability to resolve them dynamically. 
This live benchmark would serve as a versatile test bed for various experiments and the development of new solutions to improve existing management systems. 

\section{Conclusion}

In conclusion, this study has presented a novel approach to advancing the Vision of Autonomic Computing by leveraging Large Language Models within a hierarchical multi-agent framework for microservice management. Through the introduction of a five-level taxonomy for autonomous service maintenance and the development of an online evaluation benchmark based on the Sock Shop microservice demo, we systematically assessed the performance of our proposed system. Our findings reveal that the LLM-based multi-agent framework achieves Level 3 autonomy, effectively handling detection and resolution of specific issues, although there remains room for enhancement in areas such as root cause analysis and mitigation. 
This work not only demonstrates the potential of LLMs to significantly contribute towards realizing ACV but also sets the stage for future research to address the remaining challenges and push the boundaries of self-managing computing systems further.

%%
%% The acknowledgments section is defined using the "acks" environment
%% (and NOT an unnumbered section). This ensures the proper
%% identification of the section in the article metadata, and the
%% consistent spelling of the heading.
\newpage
\begin{acks}
We thank Lingxiang Hu, Shurun Yuan and Haiyang Ding for early participation of this work. We also appreciate David Liu for the insightful discussions and Bo Qiao for the assistance in setting up the experiment server.
\end{acks}

%%
%% The next two lines define the bibliography style to be used, and
%% the bibliography file.
\bibliographystyle{ACM-Reference-Format}
\bibliography{main}

%%
%% If your work has an appendix, this is the place to put it.
\appendix
\newpage
\section{Experiment Setup Details}

\subsection{Traffic Levels Definition with Locust}
We employed Locust to create traffic for the microservice components, emulating real users operations. We outlined four traffic modes that exert varying levels of pressure on the system. The corresponding configurations can be found in Table \ref{tb:traffic}.

\begin{table}[t]
  \centering
  \begin{tabular}{c | c | c | c  }
  \hline
  \hline
  & Traffic Level & Users & Spawn Rate \\
  \hline
  \multirow{4}{*}{\texttt{Catalogue}} & Light Traffic & 20 & 20 \\
  & Moderate Traffic & 50 & 50 \\
  & Heavy Traffic & 80 & 80 \\
  & Rising Traffic & 100 & 1 \\
  \hline
  \multirow{4}{*}{\texttt{Front-end}} & Light Traffic & 20 & 20 \\
  & Moderate Traffic & 40 & 40 \\
  & Heavy Traffic & 80 & 80 \\
  & Rising Traffic & 100 & 1 \\
  \hline
  \hline
  \end{tabular}
  \caption{Details of the defined traffic levels using \texttt{Locust}.}
  \label{tb:traffic}
\end{table}

\subsection{Detailed Description for L3/L4/L5 Tasks}
\label{sec:task_description_L3_4_5}

\begin{table*}
    \centering
    \footnotesize
    \begin{tabular}{p{15cm}}
    \toprule
    This task is a regular check for the state of your microservice component. If a microservice component does not meet any criteria for a healthy state, you need to analyze the root cause and take corrective actions to recover it. \\
    \\
    \# The healthy state of the microservice component is defined as follows: \\
      - The microservice component is running healthily with READY state. \\
      - The CPU/Memory usage of the microservice component is below than 50\% of allocated/limit resource. \\
      - The P99 latency of the microservice component is in a stable range, with no big fluctuations, and the average P99 latency is below 200ms. \\
    \\
     \# Follow the steps below to complete the task: \\
      1. Check the state of the microservice component to see if the healthy state is maintained. \\
      2. If NONE of the above healthy state criteria is violated, you can report the result directly. \\
      3. If ANY of the above healthy state criteria is violated, you should follow the steps below: \\
        - Analyze the root cause of the unhealthy state. You should try your best to identify the root cause. If the root cause is hard to identify, you still need to take corrective actions to temporarily mitigate the issue. \\
        - Take corrective actions to recover the unhealthy state. \\
        - Confirm the healthy state is restored. \\
        - You are allowed to repeat the above steps a few times until the healthy state is restored. \\
        - If the issue persists after a few attempts, you should report the issue by calling \texttt{\`}report\_result\texttt{\`} function. Note that NEVER report the issue before you have tried to recover the unhealthy state. \\
    \bottomrule
    \end{tabular}
    \caption{Detailed description for L3/L4/L5 tasks.}
    \label{tab:task-description-L3_4_5}
\end{table*}

To assess the potential of LLMs to achieve L3/L4/L5 levels of autonomy, we established a standardized task that aligns with the previously defined SLOs. We outlined clear criteria for determining the healthy state of a microservice component, and provided guidance to the agent on attaining L3/L4/L5. Additional information can be found in Table \ref{tab:task-description-L3_4_5}.

\section{Additional Experiment Results}

Here we provide more detailed results corresponding to experiments of evaluating the L3/L4/L5 tasks applied to the low-level autonomic agent \texttt{Catalogue} (Table~\ref{tb:ind_L3_4_5}), the L1/L2 tasks applied to the high-level group manager (Table~\ref{tb:group_L1_2}), and the L3/L4/L5 tasks applied to the high-level group manager (Table~\ref{tb:group_L3_4_5}).

\begin{table*}[t]
  \centering
  \small
  \begin{tabular}{c | c | c | c  c  c  c  c }
  \hline
  \hline
  \multirow{2}{*}{\makecell{Injected \\Faults}} & Trial & Eval & \multirow{2}{*}{Steps}  & \multirow{2}{*}{\makecell{Is Issue Present \\ before Eval?}} & \textbf{L3} & \textbf{L4} & \textbf{L5} \\
  & Index & Index  & &   & (Is Correct Judgement?) & (Is RCA?) & (Is Mitigated?)  \\
  \hline
  \multirow{9}{*}{\makecell{Pod \\ Failure}} & \multirow{3}{*}{1} & 1 & 7 & 1  & 1 & \makecell{misguided by message \\ in deployment description}
  & \makecell{accidentally resolve the issue \\ but introduce a new issue}\\ 
  & & 2 & 17 & 1  & 1 & 1 & \makecell{Unable to restore back \\the correct setting}\\
  & & 3 & 15 & 1  & 1 & 1 & \makecell{Unable to restore back \\the correct setting} \\
  \cline{2-8}
& \multirow{3}{*}{2} & 1 & 9 & 1  & 1 & 1  & 1 \\ 
  & & 2 & 11 & 0  & 1 & - & - \\
  & & 3 & 11 & 0  & \makecell{misguided by output \\ from wrong operation}
 & - & - \\
  \cline{2-8}
& \multirow{3}{*}{3} & 1 & 19 & 1  & 1 & 1  & 1 \\ 
  & & 2 & 4 & 0  & 1 & - & - \\
  & & 3  & 5 & 0   & 1 & - & - \\
  \hline
  \multirow{9}{*}{\makecell{CPU \\Stress}} & \multirow{3}{*}{1} & 1 & 7 & 1  & \makecell{hallucination during \\numerical comparison}
 & -  & - \\ 
  & & 2 & 9 & 1  & 1 & 1 & \makecell{wrong action by \\reducing CPU request}\\
  & & 3 & 9 & 1  & 1 & no check on CPU limit
 & 1 \\
  \cline{2-8}
& \multirow{3}{*}{2} & 1 & 13 & 1  & 1 & no check on CPU limit  & \makecell{ineffective action by \\performing horizontal scaling} \\ 
  & & 2 & 11 & 1  & 1 & no check on CPU limit & \makecell{ineffective action by \\performing horizontal scaling} \\
  & & 3 & 11 & 1  & 1 & no check on CPU limit & \makecell{ineffective action by \\performing horizontal scaling} \\
  \cline{2-8}
& \multirow{3}{*}{3} & 1 & 4 & 1  & \makecell{hallucination during \\numerical comparison} & -  & - \\ 
  & & 2 & 9 & 1  & 1 & no check on CPU limit & \makecell{ineffective action by \\performing horizontal scaling} \\
  & & 3 & 1 & 1  & task omission & - & - \\
 \hline
  \multirow{9}{*}{\makecell{Rising \\Traffic}} & \multirow{3}{*}{1} & 1 & 4 & 0  & 1 & -  & - \\ 
  & & 2 & 4 & 0  & 1 & - & - \\ 
  & & 3 & 1 & 1 & task omission & - & - \\
  \cline{2-8}
& \multirow{3}{*}{2} & 1 & 18 & 0  & 1 & -  & - \\ 
  & & 2 & 7 & 0  & 1 & -  & - \\
  & & 3 & 7 & 0  & 1 & - & - \\
  \cline{2-8}
& \multirow{3}{*}{3} & 1 & 13 & 1  & 1 & 1  & \makecell{hallucination during \\unit conversion} \\ 
  & & 2 & 7 & 1  & \makecell{hallucination during \\numerical comparison} & - & - \\
  & & 3 & 4 & 1  & \makecell{hallucination during \\numerical comparison} & - & - \\
\hline
\multicolumn{3}{c|}{\textsc{Average}} & \textbf{8.8} & All & \textbf{0.74 (20/27)} & \textbf{0.5 (6/12)} & \textbf{0.25 (3/12)} \\
\hline
\hline
\multicolumn{8}{l}{\emph{Note1:} ``Eval Index'' denotes the three consecutative evaluations in an experiment run.} \\
\multicolumn{8}{l}{\emph{Note2:} ``Is RCA'', ``Is Correct Judgement'' and ``Is Mitigated'' with text denote agent failed to achieve the goal with corresponding reasons in cells.} \\
\multicolumn{8}{l}{\emph{Note3:} ``Is Correct Judgement'' denotes that whether the agent can correctly identify the issue.} \\
\multicolumn{8}{l}{\emph{Note4:} ``Is RCA'' indicates whether the agent can accurately identify the root cause of the issue.} \\
\multicolumn{8}{l}{\emph{Note5:} ``Is Mitigated'' indicates whether the agent can address the identified issue using any available methods.} \\
  \end{tabular}
  \caption{Detailed results for L3/L4/L5 tasks applied to the low-level autonomic agent of \texttt{Catalogue}.}
  \label{tb:ind_L3_4_5}
  
\end{table*}

\begin{table*}[t]
  \centering
  \footnotesize
  \begin{tabular}{c | c | c  c  c  c  }
  \hline
  \hline
  \makecell{Task \\ Name} & \makecell{Round \\Index} & Steps & \makecell{Is Correct \\ Task Assignment?} & \makecell{Is Sub-task Finished \\ by \texttt{Catalogue}?} & \makecell{Is Sub-task Finished \\ by \texttt{Front-end}?} \\
  \hline
  \multirow{9}{*}{\makecell{Latency \\Reduction-Cat.}} & 1 & [3,5,-] & 1 & 1 & -\\
  & 2 & [2,8,-] & 1 & 1 & - \\
  & 3 & [1,-,-] & 1 & - & - \\
  \cline{2-6}
  & 1 & [2,5,-] & 1 & 1 & -\\
  & 2 & [3,11,-] & 1 & 0 & - \\
  & 3 & [3,10,-] & 1 & 1 & - \\
  & 4 & [1,-,-] & 1 & - & - \\  
  \cline{2-6}
  & 1 & [2,5,-] & 1 & 1 & -\\
  & 2 & [1,-,-] & 1 & - & - \\
  \hline
  \multirow{12}{*}{\makecell{CPU \\Reduction-Cat.}} & 1 & [2,5,-] & 1 & 1 & -\\
  & 2 & [2,9,-] & 1 & \makecell{wrong action leads to\\faked task completion} & - \\
  & 3 & [1,-,-] & \makecell{manager terminate task\\based on \texttt{Catalogue}'s report} & - & - \\
  \cline{2-6}
  & 1 & [2,5,-] & 1 & 1 & -\\
  & 2 & [2,17,-] & 1 & 1 & - \\
  & 3 & [1,-,-] & 1 & - & - \\  
  \cline{2-6}
  & 1 & [2,4,-] & 1 & 1 & -\\
  & 2 & [3,8,-] & 1 & \makecell{wrong action leads to\\faked task completion} & -\\
  & 3 & [2,-,-] & \makecell{manager terminate task\\based on \texttt{Catalogue}'s report} & - & - \\
  \hline
  \multirow{9}{*}{\makecell{Latency \\Reduction-Group}} & 1 & [2,5,5] & 1 & 1 & 1\\
  & 2 & [2,9,5] & 1 & 1 & 1 \\
  & 3 & [1,-,-] & 1 & - & - \\
  \cline{2-6}
  & 1 & [2,5,5] & 1 & 1 & 1\\
  & 2 & [2,10,11] & 1 & 1 & 1 \\
  & 3 & [1,-,-] & 1 & - & - \\  
  \cline{2-6}
  & 1 & [2,5,5] & 1 & 1 & 1\\
  & 2 & [2,11,11] & \makecell{incorrectly require the P99 \\ latency of each below 400 ms} & 1 & 1 \\
  & 3 & [1,-,-] & 1 & - & - \\  
  \hline
  \multirow{16}{*}{\makecell{CPU \\Reduction-Group}} & 1 & [2,4,5] & 1 & 1 & 1\\
  & 2 & [3,11,9] & \makecell{no check on CPU limit\\leads to wrong tasks} & \makecell{wrong action by adjusting\\CPU requests; treat task as\\completed without post-check} & \makecell{scale up to 2 replicas but\\no enough for completing task} \\
  & 3 & [1,-,-] & \makecell{manager terminate task based on \\incorrect low-level reports} & - & - \\
  \cline{2-6}
  & 1 & [2,5,5] & 1 & 1 & 1\\
  & 2 & [3,9,11] & 1 & \makecell{scale up to 2 replicas but\\no enough for completing task;\\hallucination on report} & \makecell{scale up to 3 replicas but not\\enough (slightly above the limit)} \\
  & 3 & [1,-,-] & 1 & - & - \\
  \cline{2-6}
  & 1 & [2,4,5] & 1 & 1 & 1\\
  & 2 & [2,13,9] & 1 & \makecell{wrong action by\\reducing CPU request} & \makecell{wrong action by\\reducing CPU request}\\
  & 3 & [3,6,8] & 1 & \makecell{no action taken} & \makecell{scale up to 2 replicas but not enough;\\hallucination on report} \\
  & 4 & [3,6,-] & 1 & 1 & - \\
  & 5 & [1,-,-] & 1 & - & - \\
  \hline
\multicolumn{2}{c|}{\textsc{Average}} & \textbf{[1.9, 7.5, 7.2]} & \textbf{0.87 (34/39)} & \textbf{0.77 (20/26)} & \textbf{0.69 (9/13)} \\
  \hline
  \hline
  \multicolumn{6}{l}{\emph{Note1:} The elements of each 3-tuple list in ``Steps'' represent the number of steps used by the high-level group manager, \texttt{Catalogue} and \texttt{Front-end}, respectively.} \\
  \multicolumn{6}{l}{\emph{Note2:} ``Is Correct Task Assignment'' and ``Is Sub-task Finished by \emph{X}'' with text denote agent failed to achieve the goal with corresponding reasons in cells.} \\
  \multicolumn{6}{l}{\emph{Note3:} ``Is Correct Task Assignment'' denotes whether the high-level group manager decomposed tasks/assigned sub-tasks correctly.} \\
  \multicolumn{6}{l}{\emph{Note4:} ``Is Sub-task Finished by \emph{X}'' denotes  whether low-level autonomic agent solved the sub-tasks correctly.} \\
  \end{tabular}
  \caption{Detailed results for L2 tasks applied to the high-level group manager.}
  \label{tb:group_L1_2}
\end{table*}

\begin{table*}[t]
  \centering
  \footnotesize
  \begin{tabular}{c | c | c | c  c  c  c  c  }
  \hline
  \hline
  \makecell{Injected\\Faults} & Round & \makecell{Is Issue Present\\Before Round?} & \makecell{Steps} & \makecell{Is Correct\\Task Assignment?} & \makecell{\textbf{L3}\\(Is Correct\\judgement?)} & \makecell{\textbf{L4}\\(Is RCA?)} & \makecell{\textbf{L5}\\(Is Mitigated?)} \\
  \hline
  \multirow{7}{*}{\makecell{Pod \\ Failure}} & 1 & [1,0] & [2,19,6] & 1 & [1,-] & [1,-] & [1,-]\\ 
  & 2 & {\color{black}{[0, 0]}} & [1,-,-] & 1 & - & - & - \\
  \cline{2-8}
   & 1 & [1,0] & [2,9,9] & 1 & [1,-] & [-,-] & [-,-]\\ 
  & 2 & [1,0] & [3,7,-] & 1 & [1,-] & \makecell{[misguided by \\hallucinated logs,-]} & [1,-]\\ 
  & 3 & {\color{black}{[0, 0]}} & [1,-,-] & 1 & - & - & - \\
  \cline{2-8}
  & 1 & [1,0] & [2,20,7] & 1 & [1,-] & \makecell{[misguided by\\logs in the output,-]} & [1,-]\\ 
  & 2 & {\color{black}{[0, 0]}} & [1,-,-] & 1 & - & - & - \\
  \hline
  \multirow{13}{*}{\makecell{CPU \\ Stress}} & 1 & [1,1] & [2,6,9] & 1 & - & - & -\\ 
  & 2 & [1,1] & [3,10,8] & 1 & [1, task omission] & [1, task omission] & \makecell{[1, increase resource\\allocation but not enough]}\\
  & 3 & [0,1] & [3,-,8] & 1 & [-,1] & [-, task omission] & \makecell{[-, yes but \\ report incorrectly]}\\ 
  & 4 & [0,0] & [3,-,8] & 1 & [-, 1] & - & -\\ 
  & 5 & {\color{black}{[0, 0]}} & [1,-,-] & 1 & - & - & - \\
  \cline{2-8}
  & 1 & [1,1] & [2,9,9] & 1 & - & - & -\\ 
  & 2 & [1,1] & [3,11,15] & omit analysis task & - & - & \makecell{[increase resource allocation but\\didn't apply, scale up to 2 replicas\\and increase resource allocation\\but not enough]}\\
  & 3 & [1,1] & [3,11,7] & omit analysis task & - & - & [1,1]\\ 
  & 4 & {\color{black}{[0, 0]}} & [1,-,-] & 1 & - & - & - \\
  \cline{2-8}
  & 1 & [1,1] & [2,6,9] & 1 & - & - & -\\ 
  & 2 & [1,1] & [3,7,11] & omit analysis task & - & - & \makecell{[increase resource allocation\\but not apply, 1]}\\
  & 3 & [1,0] & [3,8,-] & omit analysis task & - & - & [1,-]\\ 
  & 4 & {\color{black}{[0, 0]}} & [1,-,-] & 1 & - & - & - \\
  \hline
  \multirow{13}{*}{\makecell{Rising \\Traffic}} & 1 & [1,1] & [2,6,9] & 1 & - & - & -\\ 
  & 2 & [1,1] & [3,15,11] & omit analysis task & - & - & \makecell{[increase CPU request beyond the limit\\resulting in failed apply, increase resource\\limit but not apply successfully]}\\
  & 3 & [1,1] & [3,8,8] & omit analysis task & - & - & \makecell{[yes but still thought the task was incomplete\\due to hallucination on numerical comparison,\\increase CPU limit but decrease memory limit\\resulting in task failure]}\\
  & 4 & {\color{black}{[1, 1]}} & [1,-,-] & \makecell{give up by manager} & - & - & - \\
  \cline{2-8}
  & 1 & [1,1] & [2,9,9] & 1 & - & - & -\\ 
  & 2 & {\color{black}{[1, 1]}} & [1,-,-] & \makecell{termination due to\\hallucination on\\CPU limit by manager} & - & - & - \\
  \cline{2-8}
  & 1 & [1,1] & [2,8,6] & 1 & - & - & -\\ 
  & 2 & [1,1] & [3,9,9] & 1 & [1,1] & [1,1] & -\\
  & 3 & [1,1] & [3,2,6] & 1 & - & - & \makecell{[no action taken, \\yes but omit report]}\\ 
  & 4 & {\color{black}{[1, 0]}} & [1,-,-] & \makecell{hang up on awaiting\\low-level responses} & - & - & - \\
  \hline
\multicolumn{3}{c|}{\textsc{Average}} & \textbf{[2.1, 9.5, 8.6]} & \textbf{0.7 (21/30)} & \textbf{0.9 (9/10)} & \textbf{0.5 (4/8)} & \textbf{0.42 (8/19)}\\
\hline
\hline
\multicolumn{8}{l}{\emph{Note1:} The elements of each 2-tuple list in ``Is Issue Present Before Round'' indicate whether there exist issues for \texttt{Catalogue} and \texttt{Front-end}, respectively.} \\
  \multicolumn{8}{l}{\emph{Note2:} The elements of each 3-tuple list in ``Steps'' represent the number of steps used by the high-level group manager, \texttt{Catalogue} and \texttt{Front-end}, respectively.} \\
  \multicolumn{8}{l}{\emph{Note3:} The elements of each 2-tuple list in the L3/L4/L5 columns correspond to the results for \texttt{Catalogue} and \texttt{Front-end}, respectively.} \\
  \end{tabular}
  \caption{Detailed results for L3/L4/L5 tasks applied to the high-level group manager.}
  \label{tb:group_L3_4_5}
\end{table*}

\section{Prompts}
\label{sec:prompt}

The specific system prompts used for the low-level autonomic agents and high-level group manager are presented in Table \ref{tab:low-level-agent-prompt} and \ref{tab:high-level-manager-prompt}, respectively.

\begin{table*}[htbp]
    \centering
    \footnotesize
    \begin{tabular}{p{15cm}}
    \toprule
    - You are a Kubernetes component maintainer named \textcolor{blue}{"\{\{service\_name\}\}"} with k8s manager role to ensure that the microservice component \textcolor{blue}{"\{\{service\_name\}\}"} is running normally and healthily. \\
    - You are mainly responsible for two types of tasks: answering maintenance-related inquiries (e.g., what is current component status/resource usage) and providing instructions to achieve maintenance requests (e.g., reduce latency to 10ms, update the image version). \\
    - You are provided with basic information of the component in section \texttt{\`}Component Information\texttt{\`}. (e.g., description/namespace/deployment artifacts) \\
    - Use available tools to help you analyze the component status and perform necessary maintenance operations. (e.g., Kubernetes, Prometheus, Tool Functions) \\
    - Before starting work, you should read all the information in \texttt{\`}Component Information\texttt{\`} and \texttt{\`}Tools Information\texttt{\`} sections to better understand the component you are maintaining and available tools you can leverage.
    - Follow the \texttt{\`}Instructions\texttt{\`} section to take actions. \\
    \\
    \# Component Information: \\
    - The description of the component is \textcolor{blue}{"\{\{service\_description\}\}"}. \\
    - The component is under the namespace of \textcolor{blue}{\{\{namespace\}\}}. \\
    - This component is deployed as a k8s service using YAML files. \\
    - The deployment YAML file is located at \textcolor{blue}{\texttt{\`}\{\{deploy\_YAML\_fp\}\}\texttt{\`}}. \\
    - The service YAML file is located at \textcolor{blue}{\texttt{\`}\{\{service\_YAML\_fp\}\}\texttt{\`}}. \\
    - Downstream dependency: the current service depends on the following list of services: \textcolor{blue}{\{\{downstream\_services\}\}} \\
    - Upstream dependency: the following list of services depend on the current service: \textcolor{blue}{\{\{upstream\_services\}\}} \\
    \\
    \# Tools Information: \\
    \#\# Kubernetes \\
    - You have the full access to the internal network of k8s cluster and you can run commands with "kubectl" command to manage the cluster. \\
    - Kubernetes Metrics Server is running by default in the cluster. \\
    - You can use \texttt{\`}kubectl top\texttt{\`} command to get some metrics of the service. \\
    \\
    \#\# Prometheus \\
        - Prometheus server is running at \textcolor{blue}{\{\{prometheus\_url\}\}}. \\
    \\
    \# Instructions \\
    \#\# Overall flow of action taking \\
    - ALWAYS follow the flow of action taking: Understand Task -> Make Plan -> Execute Plan -> Report Result -> Terminate. Note that Make Plan and Execute Plan are iterative. \\
    - The overall action taking process is to first make detailed and executable plan in steps, then take actions following the steps to solve the task. \\
    - After taking actions, double confirm if the changes does take effect and meet the goal of the task. \\
    - Before reporting result, you should try a few times to achieve the request. \\
    - Whether the task is successful or failed, ALWAYS remember to report results by calling the \texttt{\`}report\_result\texttt{\`} function, otherwise, the task assigner will not be able to get the result you have done, and the task will be considered as not completed. \\
    - Terminate the task after reporting the result. \\
    \\
    \#\# Instructions on how to make a plan \\
    - Read the task carefully and understand what you need to do. Double check the task and determine if it is feasible or reasonable. \\
    - Break down the task into a series of executable steps, each executable step should be clear. \\
    - Output your plan in the following format: \\
    Task: <Task description> \\
        Steps: \\
            1. <Step 1 description> \\
            2. <Step 2 description> \\
            3. <Step 3 description> \\
            4. <Report Result by calling \texttt{\`}report\_result\texttt{\`}> \\
            5. <Terminate> \\
        - ALWAYS explicitly output the above plan, otherwise, some steps (e.g., report result) may be missed. \\
        - Use plain text to describe the steps, DO NOT include code or command in the plan. \\
        - If the task is not solved by the initial plan, you should modify the plan and try a few more times. \\
    \\
    \#\# Instructions on how to output code/command for each executable step \\
    - Output code or commands for one step at a time. DO NOT combine code or commands for multiple steps. \\
        - Each step should be given in terms of executable codes or commands. Do NOT write code or command that is irrelevant to the task. \\
        - Use \texttt{\`}python\texttt{\`} for code, \texttt{\`}bash\texttt{\`} for command line. Do NOT output other type code blocks, especially for YAML. You should try to write a python script to generate the YAML file or modify an existing one. \\
        - You can run \texttt{\`}cat\texttt{\`} command to read the content of the file, and then output the content in the code block. \\
        - You are allowed to modify the code through a python code snippet if the task requires code modification. \\
        - When you are writing commands or code, DO NOT leave placeholders in the commands or code. If there are placeholders, you should replace them with the actual values. (e.g., pod name, container name, namespace) \\
        - ALWAYS wait for a while after taking actions that will cause changes to the system, and then check if the issue is fixed. For example, use \texttt{\`}sleep 120;\texttt{\`} command to wait for 120s. \\
        - Code/command blocks should be wrapped by \texttt{\`}\texttt{\`}\texttt{\`} (three backticks), not in plain text or markdown. \\
        - Example: for a python code snippet, the code output could look like: \\
            \texttt{\`}\texttt{\`}\texttt{\`}python \\
            <your code> \\
            \texttt{\`}\texttt{\`}\texttt{\`} \\
    \\
    \#\# Instructions on how to terminate the task \\
        - When the task is completed, ALWAYS output \texttt{\`}TERMINATE\texttt{\`} (i.e., no other content) to end the task. \\
        - Do **NOT** output \texttt{\`}TERMINATE\texttt{\`} before the task is completed; otherwise, the task will be ended prematurely.\\
    \\
    
    \textbf{CONTINUE ON THE NEXT PAGE} \\
    
    \bottomrule
    \end{tabular}
    \caption{Prompt for low-level autonomic agents}
    \label{tab:low-level-agent-prompt}
\end{table*}

\begin{table*}[htbp]
    \centering
    \footnotesize
    \begin{tabular}{p{15cm}}
    \toprule
    \#\# Additional instruction on checking logs \\
        - Only retrieve the latest 20 lines of logs. \\
        - Only focus on the relevant logs that are related to issues under investigation. \\
    \\
    \#\# Additional instructions for kubernetes \\
        - The actual pod/container/service name may be different from the provided one. You need to find the actual name by yourself. \\
        - NEVER **output** and **run** commands (e.g., \texttt{\`}kubectl get ... -w\texttt{\`}, \texttt{\`}kubectl port-forward\texttt{\`}, \texttt{\`}kubectl edit\texttt{\`} command) that will cause obstruction. \\
        - No output do NOT mean no issues and it could be because the command is wrong (e.g., wrong parameters/arguments) \\
    \\
    \#\# Additional Instructions for Prometheus under current environment \\
        - You can write Python code by sending query in Prometheus Query Language (PromQL) to Prometheus server to get the metrics you need. \\
        - Retrieve metrics by following steps: \\
        - Choose the right metric name and labels you need to query, you should use only one metric name in one query. \\
        - Available metrics: \\
            1. request\_duration\_seconds\_count: for query per second (QPS) metric. \\
            2. request\_duration\_seconds\_bucket: for lantency (in seconds) metric. \\
            - Available labels: \\
            1. name: the service name. \\
            2. status\_code: the status code of the request.
            3. route: the route of the request. \\
        - Follow the document in the \texttt{\`}Tool Functions\texttt{\`} section to query the metrics you need. \\
        - Use the tool function to query Prometheus server and get the metrics you need. \\
        - Below is a sample python code snippet to query QPS from Prometheus server: \\
    \begin{verbatim}
    ```python
    # Import tools function from file first
    from src.agent.tool_functions_for_agent import query_prometheus
    promQL = 'sum(rate(request_duration_seconds_count{name="catalogue",status_code=~"2..",route!="metrics"}[1m]))'
    duration = '2m'
    step = '1m'
    result = query_prometheus(promQL, duration=duration, step=step)
    print(result)
    ```
    \end{verbatim}
    - Below is a sample python code snippet to query P99 latency from Prometheus server: \\
    \begin{verbatim}
    ```python
    # Import tools function from file first
    from src.agent.tool_functions_for_agent import query_prometheus
    promQL = 'histogram_quantile(0.99, sum(rate(request_duration_seconds_bucket{name="catalogue"}[1m])) by (name, le))'
    duration = '2m'
    step = '1m'
    result = query_prometheus(promQL, duration=duration, step=step)
    print(result)
    ```
    \end{verbatim}
    - When you get empty result or nan in list, you should check if the metric name is correct and the time range is correct, correct them and try again. \\
    \\
    \#\#Additional Instructions for current microservice "sock-shop" \\
    - Using only \texttt{\`}name\texttt{\`} selector (NOT \texttt{\`}app\texttt{\`} selector) to find the pod/container/service, e.g., \texttt{\`}kubectl get pod -n \textcolor{blue}{\{\{namespace\}\}} -l name=\textcolor{blue}{\{\{service\_name\}\}}\texttt{\`} \\
    \\
    \textbf{CONTINUE ON THE NEXT PAGE} \\
    \bottomrule
    \end{tabular}
    \caption{Prompt for low-level autonomic agents (continued)}
\end{table*}

\begin{table*}[htbp]
    \centering
    \footnotesize
    \begin{tabular}{p{15cm}}
    \toprule
    \# Introduction for Tool Functions \\
    - You have access to the following tool functions. They can be accessed from the module called `\textcolor{blue}{\{model\_name\}}` by their function names.\\
    - For example, if there was a function called \texttt{\`}foo\texttt{\`} you could import it by writing \texttt{\`}from \textcolor{blue}{\{model\_name\}} import foo\texttt{\`}\\
    \\
    \begin{verbatim}
    def query_prometheus(promQL: str, **kwargs) -> list:
    """
    This function is used to query prometheus with the given promQL.
    - param promQL: str, the promQL to be executed
    - param kwargs: dict, parameters to be passed to the query, must contain one of the following: (start_time, end_time), duration
    
    return: list, result of the query
    
    Available metrics:
    1. request_duration_seconds_count: for query per second (QPS) metric.
    2. request_duration_seconds_bucket: for lantency metric.
    Available filters:
    1. name: the service name.
    2. status_code: the status code of the request.
    3. route: the route of the request.

    Note: ALWAYS call print() to report the result so that planner can get the result.

    Example: 
    >>> from src.agent.tool_functions_for_agent import query_prometheus
    >>> promQL = 'rate(request_duration_seconds_count{name="catalogue",status_code=~"2..",route!="metrics"}[1m])'
    >>> result = query_prometheus(promQL=promQL, duration='2m', step='1m')
    >>> print(result) # output the result so that planner can get it.
    [['2024-06-20 02:17:20', 0.0], ['2024-06-20 02:18:20', 0.0], ['2024-06-20 02:19:20', 0.0]]
    """
    ...
    \end{verbatim}
    \\
    \begin{verbatim}
    def report_result(component: str, message: str, message_type: Literal['ISSUE', 'RESPONSE']) -> str:
    """
    This function can help you send a message to the manager.
    - param component: str, the component name
    - param message: str, the message to be reported
    - param type: str, the type of the message, use 'ISSUE' for HEARTBEAT and 'RESPONSE' for TASK

    return: str, the result of the operation

    Note: ALWAYS call print() to report the result so that planner can get the result.

    Example:
    >>> from src.agent.tool_functions_for_agent import report_result
    >>> component = 'catalogue'
    >>> message = 'The task is completed.'
    >>> message_type = 'RESPONSE'
    >>> result = report_result(component=component, message=messages, message_type=message_type)
    >>> print(result) # output the result so that planner can get it.
    Message sent to manager.
    """
    ...
    \end{verbatim}
    
    \\
    \bottomrule
    \end{tabular}
    \caption{Prompt for low-level autonomic agents (continued)}
\end{table*}

\label{sec:prompt_high}

\begin{table*}[htbp]
    \centering
    \footnotesize
    \begin{tabular}{p{15cm}}
    \toprule

    - You are a high-level service maintainer which manages a few lower-level service maintainers in Kubernetes cluster. \\
    - Both the high-level and lower-level maintainers are intelligent service maintenance agents powered by Large Language Models. \\
    - You are responsible for two types of tasks: i) Ensure the whole Kubernetes cluster is running healthily by taking proactive actions; ii) Respond to maintainence requests or inquiries from higher-level managers or human operators. \\
    - You are provided with basic information in section \texttt{\`}Service Information\texttt{\`} about the service to maintain and the lower-level maintainers in the cluster. \\
    - Follow the \texttt{\`}Instructions\texttt{\`} section to assign tasks to lower-level maintainers and collect/analyze responses from them. \\
    \\
    \# Service Information \\
    - The cluster is a Kubernetes cluster with microservice components deployed. \\
    - The main service that you are maintaining is under the namespace of \textcolor{blue}{\{\{namespace\}\}}. \\
    - The lower-level service component maintainers you can assign task to are listed as follows: \textcolor{blue}{\{\{service\_maintainers\}\}}. \\
    \\
    \# Instructions \\
  \#\# General Instructions \\
  - Your overall workflow is: Understand Task -> Decompose Task -> Assign Task -> Collect Response -> Evaluate Response -> Terminate. Note that Decompose Task, Assign Task, Collect Response and Evaluate Response are iterative. \\
  - You manage the service ONLY through assigning tasks to lower-level service maintainers. \\
  - You are NOT allowed to directly modify the cluster components via some maintenance operations like \texttt{\`}kubectl apply\texttt{\`}, \texttt{\`}kubectl delete\texttt{\`}, etc. \\
  \\
  \#\# Instructions for task decomposition \\
  - Think carefully how the task can be done, and break down the task into steps. \\ 
  - You should make a plan to assign tasks to lower-level maintainers to get the necessary information to solve the task. \\
  - Update the task decomposition in the task description when you receive responses from lower-level maintainers. \\
  - Below is an example of how to decompose a task: \\
    Task: Reduce total latency of catalogue and front-end below 50ms. \\
    Steps: \\
      1. Get the current latency of catalogue. (Assign to catalogue maintainer) \\
      2. Get the current latency of front-end. (Assign to front-end maintainer) \\
    RESPONSE from component catalogue: The current latency of catalogue is 80ms. \\
    RESPONSE from component front-end: The current latency of front-end is 40ms. \\
    Steps: \\
      1. Reduce the latency of catalogue to below 30ms. (Assign to catalogue maintainer) \\
      2. Reduce the latency of front-end to below 20ms. (Assign to front-end maintainer) \\
    RESPONSE from component: The latency of catalogue is decreased to 30ms. \\
    RESPONSE from component: The latency of front-end is decreased to 20ms. \\
    Output: The total latency of catalogue and front-end is 50ms, \texttt{\`}TERMINATE\texttt{\`}. \\
     \\
    \#\# Instructions for task assignment \\
  - Assign tasks when you get message begin with \texttt{\`}ISSUE\texttt{\`} or \texttt{\`}TASK\texttt{\`}. (e.g., 'TASK: get CPU usage from catalogue and front-end.') \\
  - ONLY assign tasks to service maintainers listed in the \texttt{\`}Service Information\texttt{\`} section. \\
  - You can assign tasks to multiple lower-level maintainers in one step, but one maintainer ONLY receive one task at a time. \\
  - The assigned task should be declarative and high-level, and you should NOT provide specific instructions to the lower-level maintainers. For example, using 'Reduce the latency of your component below 30ms', rather than 'Reduce the latency below 30ms by scaling replica to 3'. \\
  - You can ONLY assign tasks by using the provided \texttt{\`}assign\_tasks\texttt{\`} function. \\
  - Below is the example of how to assign tasks: \\
      Input: Assign tasks to two components \\
      Your output: \\
  \begin{verbatim}
  ```python
  from src.agent.tool_functions_for_cluster_manager import assign_tasks
  components = ['<component_name1>', '<component_name2>']
  task = ['<task_description1>', '<task_description2>']
  result = assign_tasks(components, tasks)
  print(result)
  ```
      \end{verbatim}
  - ALWAYS output code blocks that are wrapped by \texttt{\`}\texttt{\`}\texttt{\`} (three backticks), not in plain text or markdown. \\
  \\
  \#\# Instructions for collecting and evaluating responses \\
  - The responses from lower-level service maintainers begin with \texttt{\`}RESPONSE\texttt{\`}. (e.g., 'RESPONSE from component: The CPU usage of catalogue is 50\%.') \\
  - You need to ensure the responses from all previously assigned tasks are collected before moving to the next step. \\
  - Reponses can be arrived in any order, and you should wait for all responses before evaluating them. \\
  - Upon receiving all responses, summarize the responses and evaluate the responses to complete the task. \\
  - If the task is not solved, reorganize the steps and assign tasks again. \\
  - If the task is solved, no actions are required, summarize the responses and output \texttt{\`}TERMINATE\texttt{\`}. \\
  \\
    \textbf{CONTINUE ON THE NEXT PAGE} \\
    \bottomrule
    \end{tabular}
    \caption{Prompt for high-level group manager}
    \label{tab:high-level-manager-prompt}
\end{table*}

\begin{table*}[htbp]
    \centering
    \footnotesize
    \begin{tabular}{p{15cm}}
    \toprule
    \# Introduction for Tool Functions \\
    - You have access to the following tool functions. They can be accessed from the module called \texttt{\`}\textcolor{blue}{\{model\_name\}}\texttt{\`} by their function names.\\
    - For example, if there was a function called \texttt{\`}foo\texttt{\`} you could import it by writing \texttt{\`}from \textcolor{blue}{\{model\_name\}} import foo\texttt{\`}\\
    \\
    \begin{verbatim}
    def assign_tasks(components: list, messages: list) -> str:
    """
    This function can help you assign the task to the specific component.
    - param component: the component to which the task is assigned
    - param message: the task message

    return: str, the result of the assignment

    Example:
    >>> from src.agent.tool_functions_for_cluster_manager import assign_tasks
    >>> components = ['catalogue', 'front-end']
    >>> messages = ['Please update the service.', 'Please restart the service.']
    >>> result = assign_tasks(components, messages)
    >>> print(result)
    Tasks assigned.
    """
    \end{verbatim}
    \\
    \bottomrule
    \end{tabular}
    \caption{Prompt for high-level group manager (continued)}
\end{table*}

\end{document}